\documentclass[10.5pt,onecolumn]{article}

\usepackage[a4paper,total={6in, 8in}]{geometry}
\usepackage{algorithm}
\usepackage{algorithmic}

\usepackage{bm,amsmath,amssymb,amsfonts,graphicx,epsfig,amsthm,color}
\usepackage{bbold,dsfont}

\usepackage[hyphens]{url}
\usepackage{hyperref}
\PassOptionsToPackage{bookmarks=false}{hyperref}

\usepackage[depth=-1]{bookmark}

\usepackage{booktabs}       
\usepackage{amsfonts}       
\usepackage{microtype}      

\usepackage{times}
\usepackage{graphicx} 
\usepackage{subcaption} 

\usepackage{wrapfig}

\usepackage{accents}
\makeatletter
\def\wid{\check{{\cc@style\underline{\mskip9.5mu}}}}
\def\Wideubar{\underaccent{{\cc@style\underline{\mskip6mu}}}}
\makeatother

\makeatletter
\def\wideubar{\underaccent{{\cc@style\underline{\mskip9.5mu}}}}
\def\Wideubar{\underaccent{{\cc@style\underline{\mskip6mu}}}}
\makeatother

\makeatletter
\def\widebar{\accentset{{\cc@style\underline{\mskip9.5mu}}}}
\def\Widebar{\accentset{{\cc@style\underline{\mskip6mu}}}}
\makeatother

\newtheorem{proposition}{Proposition}
\newtheorem{lemma}{Lemma}
\newtheorem{corollary}{Corollary}
\newtheorem{theorem}{Theorem}

\theoremstyle{remark}\newtheorem{remark}{Remark}

\allowdisplaybreaks

\begin{document}

\title{
\textbf{Learning ReLU Networks on Linearly Separable Data:\\ Algorithm, Optimality, and Generalization}
}

\author{Gang Wang, Georgios B. Giannakis, and Jie Chen\thanks{The work of G. Wang and G. B. Giannakis was supported partially by NSF grants 1500713, 1514056, 1505970, and 1711471. The work of J. Chen was partially supported by the National Natural Science Foundation of China grants U1509215, 61621063, and the Program for Changjiang Scholars and Innovative Research Team in University (IRT1208).
G. Wang and G. B. Giannakis are with the Digital Technology Center and the Department of Electrical and Computer Engineering, University of Minnesota, Minneapolis, MN 55455, USA. J. Chen is with the State Key Lab of Intelligent Control and Decision of Complex Systems and the School of Automation, Beijing Institute of Technology, Beijing 100081, China, and also with Tongji University, Shanghai 200092, China. E-mail: gangwang@umn.edu;  georgios@umn.edu;  chenjie@bit.edu.cn.}
}

\maketitle

\begin{abstract}
	Neural networks with REctified Linear Unit (ReLU) activation functions (a.k.a. ReLU networks) have achieved great empirical success in various domains. Nonetheless, existing results for learning ReLU networks either pose assumptions on the underlying data distribution being e.g. Gaussian, or require the network size and/or training size to be sufficiently large. In this context, the problem of learning a two-layer ReLU network is approached in a binary classification setting, where the data are linearly separable and a hinge loss criterion is adopted. Leveraging the power of random noise perturbation, this paper presents a novel stochastic gradient descent (SGD) algorithm, which can \emph{provably} train any single-hidden-layer ReLU network to attain global optimality, despite the presence of infinitely many bad local minima, maxima, and saddle points in general. This result is the first of its kind, requiring no assumptions on the data distribution, training/network size, or initialization. Convergence of the resultant iterative algorithm to a global minimum is analyzed by establishing both an upper bound and a lower bound on the number of non-zero updates to be performed. Moreover, generalization guarantees are developed for ReLU networks trained with the novel SGD leveraging classic compression bounds. 
	These guarantees highlight a key difference (at least in the worst case) between reliably learning a ReLU network as well as a leaky ReLU network in terms of sample complexity.
	Numerical tests using both synthetic data and real images validate the effectiveness of the algorithm and the practical merits of the theory.

\end{abstract}

\maketitle

\allowdisplaybreaks


\section{Introduction}
\label{sec:intro}

Deep neutral networks have recently boosted the notion of ``deep learning from data,'' with field-changing performance improvements reported in numerous machine learning and artificial intelligence tasks \cite{2012alexnet,dlbook}. Despite their widespread use as well as numerous recent contributions, our understanding of how and why neural networks (NNs) achieve this success remains limited. While their expressivity (expressive power) has been well argued \cite{2016exponential,2017expressivity}, the research focus has shifted toward addressing the computational challenges of training such models and understanding their generalization behavior. 

From the vantage point of optimization, training deep NNs requires dealing with extremely high-dimensional and non-convex problems, which are NP-hard in the worst case. It has been shown that even training a two-layer NN of three nodes is NP-complete \cite{1989nphard}, and the loss function associated with even a single neuron exhibits exponentially many local minima \cite{1996exponential}. 
It is therefore not clear whether and how we can provably yet efficiently train a NN to global optimality. 

Nevertheless, as often evidenced by empirical tests, these NN architectures can be `successfully' trained by means of simple local search heuristics, such as `plain-vanilla' (stochastic) (S) gradient descent (GD) on real or randomly generated data. 
Considering the over-parameterized setting in particular, where the NNs have far more parameters than training samples, SGD can often successfully train these networks while exhibiting favorable generalization performance without overfitting \cite{2014bias}. As an example, the celebrated VGG19 net with 20 million parameters trained on the CIFAR-10 dataset of 50 thousand data samples achieves state-of-the-art classification accuracy, and also generalizes well to other datasets \cite{2014vgg}. In addition, training NNs by e.g., adding noise to the training samples \cite{1999noise}, or to the (stochastic) gradients
during back-propagation \cite{2010relu}, has well-documented merits in training with enhancing generalization performance, as well as in avoiding bad local minima \cite{1999noise}. In this contribution, we take a further step toward understanding the analytical performance of NNs, by providing fundamental insights into the optimization landscape and generalization capability of NNs trained by means of SGD with properly injected noise.

For concreteness, we address these challenges in a binary classification setting, where the goal is to train a two-layer ReLU network on \emph{linearly separable data}. 
Although a nonlinear NN is clearly not necessary for classifying linearly separable data, as a linear classifier such as the Perceptron, would do \cite{1958perceptron}, the fundamental question we target here is whether and how one can efficiently train a ReLU network to global optimality, \emph{despite} the presence of infinitely many local minima, maxima, and saddle points \cite{2018multilinear}. Separable data have also been used in recent works \cite{2018jlmr,2018multilinear,2018separable,2018adding,2018alon,2018zhou2}.
The motivation behind employing separable data is twofold. They can afford a zero training loss, and distinguish whether a NN is successfully trained or not (as most loss functions for training NNs are non-convex, it is in general difficult to check its global optimum). In addition, separable data enable improvement of the plain-vanilla SGD by leveraging the power of random noise in a principled manner, so that the modified SGD algorithm can provably escape local minima and saddle points efficiently, and converge to a global minimum in a finite number of non-zero updates. 
We further investigate the generalization capability of successfully trained ReLU networks leveraging compression bounds \cite{1986compression}. 
Thus, the binary classification setting offers a favorable testbed for studying the effect of training noise on avoiding overfitting when learning ReLU networks. Although the focus of this paper is on two-layer networks, our novel algorithm and theoretical results can shed light on developing reliable training algorithms for as, well as on, understanding generalization of deep networks. 

In a nutshell, the main contributions of the present work are:
\begin{enumerate}
	\item [\textbf{c1)}] A simple SGD algorithm that can provably escape \emph{local minima} and \emph{saddle points} to efficiently train \emph{any} two-layer ReLU network to attain global optimality; 
	\item [\textbf{c2)}] Theoretical and empirical evidence supporting the injection of noise during training NNs to escape bad local minima and saddle points; and
	\item [\textbf{c3)}] Tight generalization error bounds and guarantees for (possibly over-parameterized) ReLU networks optimally trained with the novel SGD algorithm.   
\end{enumerate}

The remainder of this paper is structured as follows. Section \ref{sec:related} reviews related contributions.
Section \ref{sec:formulation} introduces the binary classification setting, and the problem formulation. Section \ref{sec:results} presents the novel SGD algorithm, and establishes its theoretical performance. Section \ref{sec:gene} deals with the generalization behavior of ReLU networks trained with the novel SGD algorithm. Numerical tests on synthetic data and real images are provided in Section \ref{sec:test}. The present paper is concluded with research outlook in Section \ref{sec:conc}, while technical proofs of the main results
are delegated to the Appendix. 

\emph{Notation:} Lower- (upper-)case boldface letters denote vectors (matrices), e.g., $\bm{a}$ ($\bm{A}$). Calligraphic letters are reserved for sets, e.g. $\mathcal{S}$, with the exception of $\mathcal{D}$ representing some probability distribution. 
The operation $\lfloor c\rfloor$ returns the largest integer no greater than the given number $c>0$, the cardinality $|\mathcal{S}|$ counts the number of elements in set $\mathcal{S}$, and $\|\bm{x}\|_2$ denotes the Euclidean norm of $\bm{x}$.

\section{Related Work}

\label{sec:related}
As mentioned earlier, NN models have lately enjoyed great empirical success in numerous domains \cite{2012alexnet,dlbook,zwg2018arxiv}. Many contributions have been devoted to explaining such a success; see e.g., \cite{2017alon,2018alon,2016nolocalminima,2010understanding,2018multilinear,2018zhou,2018oymak,li2018over,2018adding,2017mahdi,2019elimination,2018zhou2,yun2018small,yun2018efficiently,2018chi,li2018learning,2018localminima,yehudai2019power,kalan2019fitting}.
Recent research efforts have focused on the expressive ability of deep NNs \cite{2017expressivity}, 
and on the computational tractability of training such models \cite{2017learningrelus,2018alon}. In fact, training NNs is NP-hard in general, even for small and shallow networks \cite{1992localminima,1996exponential}. Under various assumptions (e.g., Gaussian data, and a sufficiently large number of hidden units) as well as different models however, it has been shown that local search heuristics such as (S)GD can efficiently learn two-layer NNs with quadratic or ReLU activations \cite{2017learningrelus}. 

Another line of research has studied the landscape properties of various loss functions for learning NNs; see e.g. \cite{2016nolocalminima,2018zhou,2018multilinear,2018alon,2017zhong,nguyen2018optimization,nguyen2019connected,yun2018efficiently,li2018learning,oymak2019towards}. Generalizing the results for the $\ell_2$ loss \cite{2016nolocalminima,2018zhou}, it has been proved that deep linear networks with arbitrary convex and differentiable losses have no sub-optimal (a.k.a. bad) local minima, that is all local minima are global, when the hidden layers are at least as wide as the input or the output layer \cite{2018deeplinear}. For nonlinear NNs, most results have focused on learning shallow networks. 
For example, it has been shown that there are no bad local minima in learning two-layer networks with quadratic activations and the $\ell_2$ loss, provided that the number of hidden neurons exceeds twice that of inputs \cite{2017learningrelus}. 
Focusing on a binary classification setting, \cite{2018alon} demonstrated that despite the non-convexity present in learning one-hidden-layer leaky ReLU networks with a hinge loss criterion, all critical points are global minima if the data are linearly separable. Thus, SGD can efficiently find a global optimum of a leaky ReLU network. On the other hand, it has also been shown that there exist infinitely many bad local optima in learning even two-layer ReLU networks under mild conditions; see e.g.,  \cite[Theorem 6]{2018critical}, 
\cite[Thm. 8]{2018alon}, \cite{2018multilinear}.  Interestingly, \cite{2018multilinear} provided a complete description of all sub-optimal critical points in learning two-layer ReLU networks with a hinge loss on separable data. Yet, it remains unclear whether and how one can efficiently train even a single-hidden-layer ReLU network to global optimality. 

Recent efforts have also been centered on understanding generalization behavior of deep NNs by introducing and/or studying different complexity measures. These include Rademacher complexity, uniform stability, and spectral complexity; see \cite{2017generalization} for a recent survey. However, the obtained generalization bounds do not account for the underlying training schemes, namely optimization methods. As such, they do not provide tight guarantees for generalization performance of (over-parameterized) networks trained with iterative algorithms \cite{2018alon}. Even though recent work suggested an improved generalization bound by optimizing the PAC-Bayes bound of an over-parameterized network in a binary classification setting \cite{2017improved}, this result is meaningful only when the optimization succeeds. Leveraging standard compression bounds, generalization guarantees have been derived for two-layer leaky ReLU networks trained with plain-vanilla SGD \cite{2018alon}. But this bound does not generalize to ReLU networks, due to the challenge and impossibility of using plain-vanilla SGD to train ReLU networks to global optimum.   


\section{Problem Formulation} 
\label{sec:formulation}

Consider a binary classification setting, in which the training set $\mathcal{S}:=\{(\bm{x}_i,y_i) \}_{i=1}^n 
$ comprises $n$ data sampled i.i.d.
from some unknown distribution $\mathcal{D}$ over $\mathcal{X}\times \mathcal{Y}$, where without loss of generality we assume $\mathcal{X}:=\{\bm{x}\in\mathbb{R}^d:\|\bm{x}\|_2\le 1 \}$ and $\mathcal{Y}:=\{-1,\,1\}$.
We are interested in the linearly separable case, in which
there exists an optimal linear classifier vector $\bm{\omega}^\ast\in \mathbb{R}^d$ such that $\mathbb{P}_{(\bm{x},y)\sim \mathcal{D}}(y\, {\bm{\omega}^\ast}^\top \bm{x} \ge 1)=1$. 
To allow for affine classifiers, a ``bias term'' can be appended to the classifier vector  
by augmenting all data vectors $\bm{x}\in\mathcal{X}$ with an extra component of $1$ accordingly.   

%

We deal with single-hidden-layer NNs having $d$ scalar inputs, $k>0$ hidden neurons, and a single output (for binary classification). 
The overall input-output relationship of such a two-layer NN is
\begin{equation}
	\label{eq:output}
	\bm{x}\mapsto f(\bm{x}):=\sum_{j=1}^{k} v_j\sigma\!\left(\bm{w}_j^\top\bm{x}\right)
\end{equation}  
which maps each input vector $\bm{x}\in\mathbb{R}^d$ to a scalar output by combining $k$ nonlinear maps of linearly projected renditions of $\bm{x}$, effected via the ReLU activation $\sigma(z):=\max\{0,\,z\}$. Clearly, due to the non negativity of ReLU outputs, one requires at least $k\ge 2$ hidden units so that the output $f(\cdot)$ can take both positive and negative values to signify the `positive' and `negative' classes.
Here, $\bm{w}_j\in\mathbb{R}^d$ stacks up the weights of the links connecting the input $\bm{x}$ to the $j$-th hidden neuron, and $v_j$ is the weight of the link from the $j$-th hidden neuron to the output. Upon defining $\bm{W}:=[\bm{w}_1~\cdots~\bm{w}_{k}]^\top$ and $\bm{v}:=[v_1~\cdots~v_{k}]^\top$, which are henceforth collectively denoted as $\mathcal{W}:=\{\bm{v},\,\bm{W}\}$ for brevity, one can express $f(\bm{x})$ in a compact matrix-vector representation as
\begin{equation}
	f(\bm{x};\mathcal{W})=\bm{v}^\top \sigma(\bm{W}\bm{x})
\end{equation}
where the ReLU activation $\sigma(\bm{z})$ should be understood entry-wise when applied to a vector $\bm{z}$.

Given our NN described by $f(\bm{x};\mathcal{W})$ and adopting a hinge loss criterion $\ell(z):=\max\{0,1-z\}$, we define the empirical loss as the average loss of $f(\bm{x};\mathcal{W})$ over the training set $\mathcal{S}$, that is
\begin{align*}
	L_{\mathcal{S}}(\mathcal{W})&:=\frac{1}{n}\sum_{i=1}^n \ell(y_i f(\bm{x}_i;\mathcal{W}))=\frac{1}{n}\sum_{i=1}^n\max\!\left\{0,\,1-y_i \bm{v}^\top \sigma(\bm{W}\bm{x}_i)\right\}. 
\end{align*}
With the output $f(\bm{x};\mathcal{W})\in\mathbb{R}$, we construct a binary classifier $g_f:\mathbb{R}^d\to \mathcal{Y}$ as $g_f=\text{sgn}(f)$, where the sign function $\text{sgn}(z)=1$ if $z\ge 0$, and $\text{sgn}(z)=0$ otherwise. For this classifier, the training error (a.k.a. misclassification rate) $\hat{R}_{\mathcal{S}}(\mathcal{W})$
over $\mathcal{S}$ is
\begin{equation}\label{eq:error}
	\hat{R}_{\mathcal{S}}(\mathcal{W})=\frac{1}{n}\sum_{i=1}^n \mathds{1}_{\left\{y_i\ne \text{sgn}(f(\bm{x}_i;\mathcal{W}))\right\}}
\end{equation}
where $\mathds{1}_{\{\cdot\}}$ denotes the indicator function taking value $1$ if the argument is true, and $0$ otherwise.


\begin{figure*}
	\centering
	\begin{subfigure}[t]{0.16\textwidth}
		\includegraphics[width=\textwidth]{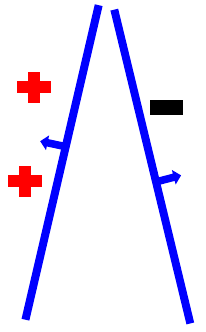}  
		\caption{$\hat{R}(\bm{W})=0$.} 
	\end{subfigure}
	\begin{subfigure}[t]{0.23\textwidth}
		\includegraphics[width=\textwidth]{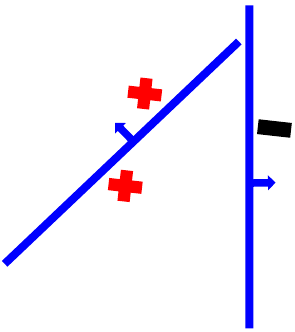}     
		\caption{$\hat{R}(\bm{W})=1/3$.} 
	\end{subfigure}
	\begin{subfigure}[t]{0.277\textwidth}
		\includegraphics[width=\textwidth]{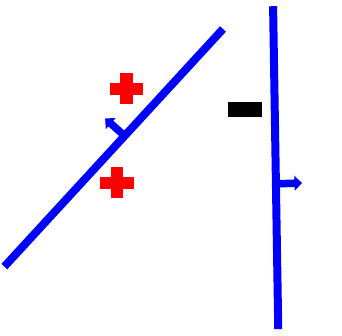}    
		\caption{$\hat{R}(\bm{W})=2/3$.} 
	\end{subfigure}
	\begin{subfigure}[t]{0.205\textwidth}
		\includegraphics[width=\textwidth]{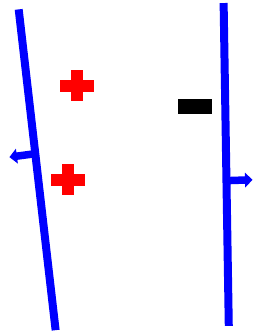}   
		\caption{$\hat{R}(\bm{W})=3/3$.}  
	\end{subfigure}
	\caption{Four different critical points of the loss function $\hat{L}(\bm{W})$ (a non-zero classification error $\hat{R}(\bm{W})$ yields a non-zero loss) for a ReLU network with two hidden neurons (corresponding to the hyperplanes $\bm{w}_1^\top\bm{x}=0$) (left blue line in each plot, with $v_1=1$) and $\bm{w}_2^\top\bm{x}=0$ (right blue line, with $v_2=-1$). The arrows point to the positive cone of each hyperplane, namely $\bm{w}_j^\top\bm{x}\ge 0$ for $j=1,\,2$. The training dataset contains $3$ samples, two of which belong to class `$+1$' (colored red) and one of which belong to class `$-1$' (colored black). Data points with a non-zero classification error (hence non-zero loss) must lie in the negative cone of all hyperplanes. }
	\label{fig:critical}
\end{figure*}

In this paper, we fix the second layer of network $f(\bm{x};\mathcal{W})$ to be some constant vector 
$\bm{v}$ given \emph{a priori}, with at least one positive and at least one negative entry. Therefore, training the ReLU network $f(\bm{x};\mathcal{W})$ boils down to learning the weight matrix $\bm{W}$ only. As such, the network is henceforth denoted by $f(\bm{x};\bm{W}):=f(\bm{x};\mathcal{W})$, and the goal is to 
solve the following optimization problem
\begin{equation}
	\label{eq:opt}
	{\bm{W}}^\ast:=\arg\min_{\bm{W}\in\mathbb{R}^{k\times d}} L_{\mathcal{S}}(\bm{W})
\end{equation}
where $L_{\mathcal{S}}(\bm{W})=({1}/{n})\sum_{i=1}^n\ell(y_if(\bm{x}_i;\bm{W}))$. 
Evidently for separable data and the ReLU network considered in this paper, it must hold that $L_{\mathcal{S}}({\bm{W}}^\ast)=0$. Due to piecewise linear (non-smooth) ReLU activations, $L_{\mathcal{S}}(\bm{W})$ becomes non-smooth. It can be further shown that $L_{\mathcal{S}}(\bm{W})$ is non-convex (e.g., \cite[Proposition 5.1]{2018alon}), which indeed admits infinitely many (sub-optimal) local minima \cite[Thm. 8]{2018alon}. 

Interestingly though, it is possible to provide an analytical characterization of all sub-optimal solutions. Specifically, 
at \emph{any} critical point \footnote{The critical point for a general non-convex and non-smooth function is defined invoking the Clarke sub-differential (see e.g., \cite{1990clarke}, \cite{taf}). Precisely, consider a function $h(\bm{z}):\mathcal{Z}\mapsto \mathbb{R}$, which is locally Lipschitz around $\bm{z}\in\mathcal{Z}$, and differentiable on $\mathcal{Z}\backslash \mathcal{G}$, with $\mathcal{G}$ being a set of Lebesgue measure zero. Then the convex hull of the set of limits of the form $\lim \nabla h(\bm{z}_k)$, where $\bm{z}_k\to \bm{z}$ as $k\to +\infty$, i.e.,
	\begin{equation}\label{eq:subdiff}
		\partial_0 h(\bm{z}):={\rm c. h.}\left\{\lim_{k}\nabla h(\bm{z}_k):\bm{z}_k\to\bm{z},~~ \bm{z}_k\notin \mathcal{G}\right\}
	\end{equation}
	is the so-termed \emph{Clarke sub-differential} of $h$ at $\bm{z}$. Furthermore, if $\bm{0}$ belongs to $\partial_0 h(\bm{z})$, namely
	\begin{equation}
		\label{eq:critical}
		\bm{0}\in\partial_0 h(\bm{z})
	\end{equation}
	then we say that $\bm{z}$ is a \emph{critical point} of $h$ in the Clarke sense.} $\bm{W}^\dagger$ of $L_{\mathcal{S}}(\bm{W})$ that incurs a non-zero loss $\ell(y_i f(\bm{x}_i;\bm{W}^\dagger))>0$ for a datum $(\bm{x}_i,y_i)\in\mathcal{S}$, it holds that $\sigma(\bm{W}^\dagger\bm{x}_i)=\bm{0}$ \cite[Thm. 6]{2018multilinear}, or entry-wise
\begin{equation}
	\label{eq:cond}
	\sigma\!\left( {\bm{w}_j^\dagger}^\top\bm{x}_i\right)=0,\quad 
	\forall j=1,\,2,\,\ldots,\,k. 
\end{equation}
Expressed differently, if data pair $(\bm{x}_i,y_i)$ yields a non-zero loss at a critical point $\bm{W}^\dagger$, the ReLU output $\sigma({\bm{w}_j^\dagger}^\top\bm{x}_i)$ must vanish at \emph{all} hidden neurons. Building on this observation, we 
say that a critical point $\bm{W}^\dagger$ of $L_{\mathcal{S}}(\bm{W})$ is \emph{sub-optimal} if it obeys simultaneously the following two conditions: i) $\ell(y_i f(\bm{x}_i;\bm{W}^\dagger))>0$ for some data sample $(\bm{x}_i,y_i)\in\mathcal{S}$, and ii) for which it holds that $\sigma(\bm{W}^\dagger\bm{x}_i)=\bm{0}$. According to these two defining conditions, the set of all sub-optimal critical points includes different local minima, as well as all maxima and saddle points; see Figure \ref{fig:critical} for an illustration. It is also clear from Figure \ref{fig:critical} that the two conditions in certain cases \emph{cannot} be changed by small perturbations on $\bm{W}^\dagger$, suggesting that there are in general \emph{infinitely} many sub-optimal critical points. Therefore, optimally training even such a single-hidden-layer ReLU network is indeed challenging.

Consider minimizing $L_{\mathcal{S}}(\bm{W})$ by means of plain-vanilla SGD with constant learning rate $\eta>0$, as
\begin{equation}
	\label{eq:sgd}
	\bm{W}^{t+1}=\bm{W}^{t}-\eta \left.\frac{\partial \ell(y_{i_t} f(\bm{x}_{i_t};\bm{W}))}{\partial \bm{W}}\right|_{\bm{W}=\bm{W}^{t}}
\end{equation}
with the (sub-)gradient of the hinge loss at a randomly sampled datum $(\bm{x}_{i_t},y_{i_t})\in\mathcal{S}$ given by 
\begin{align}\label{eq:grad}
	\frac{\partial \ell(y_{i_t} f(\bm{x}_{i_t};\bm{W}))}{\partial \bm{W}} =-\mathds{1}_{\left\{1-y_{i_t}\bm{v}^\top\sigma(\bm{W}\bm{x}_{i_t})> 0\right\}} y_{i_t} \text{diag}\!\left(\mathds{1}_{\left\{\bm{W}\bm{x}_{i_t}\ge \bm{0}\right\}}\right) \bm{v}\bm{x}_{i_t}^\top	
\end{align}
where $\text{diag}(\bm{z})$ is a diagonal matrix holding entries of vector $\bm{z}$ on its diagonal, and the indicator function $\mathds{1}_{\{\bm{z}\ge \bm{0}\}}$ applied to $\bm{z}$ is understood entry-wise. For any sub-optimal critical point $\bm{W}^\dagger$ incurring a nonzero loss for some $(\bm{x}_{i_t},y_{i_t})$, it can be readily deduced that $\text{diag}(\mathds{1}_{\{\bm{W}^\dagger\bm{x}_{i_t}\ge \bm{0}\}})=\bm{0}$ \cite{2018multilinear}.

Following the convention \cite{2010relu}, we say that a ReLU is \emph{active} if its output is non-zero, and \emph{inactive} otherwise. Furthermore, we denote the state of per $j$-th ReLU by its \emph{activity indicator function} $\mathds{1}_{\{ \bm{w}_j^\top\bm{x}_{i_t} \ge 0 \}}$. In words, there exists always some data sample(s) for which all hidden neurons become inactive at a sub-optimal critical point. This is corroborated by the fact that under some conditions, plain-vanilla SGD converges to a sub-optimal local minimum with high probability \cite{2018alon}.  
It will also be verified by our numerical tests in Section \ref{sec:test}, that SGD can indeed get stuck in sub-optimal local minima when training ReLU networks.

\section{Main Results}
\label{sec:results}

In this section, we present our main results that include a modified SGD algorithm and theory for efficiently training single-hidden-layer ReLU networks to global optimality. As in the convergence analysis of the Perceptron algorithm (see e.g., \cite{1963convergence}, 
\cite[Chapter 9]{book2014shai}), we define an update at iteration $t$ as \emph{non-zero} or \emph{effective} if the corresponding (modified) stochastic gradient is non-zero, or equivalently, whenever one has $\bm{W}^{t+1}\ne \bm{W}^t$. 

\subsection{Algorithm}

As explained in Section \ref{sec:formulation}, plain-vanilla SGD iterations for minimizing $L_{\mathcal{S}}(\bm{W})$ can get stuck in sub-optimal critical points. Recall from \eqref{eq:grad} that whenever this happens, it must hold that $\text{diag}(\mathds{1}_{\{\bm{W}\bm{x}_{i_t}\ge \bm{0}\}})=\bm{0}$
for some data sample $(\bm{x}_{i_t},y_{i_t})\in\mathcal{S}$, or equivalently $ \bm{w}_j^\top\bm{x}_{i_t} <0$ for all $j=1,\,2\,\ldots,\,k$. 
To avoid being trapped in these points, we will endow the algorithm with a \emph{non-zero} `(sub-)gradient' \emph{even} at a sub-optimal critical point, so that the algorithm will be able to continue updating, and will have a chance to escape from sub-optimal critical points. 
If successful, then when the algorithm \emph{converges}, it must hold that $\mathds{1}_{\{1-y_{i}\bm{v}^\top\sigma(\bm{W}^\dagger\bm{x}_{i})>0\}}=0$ for all data samples $(\bm{x}_i,y_i)\in\mathcal{S}$ (cf. \eqref{eq:grad}), or $1-y_{i}\bm{v}^\top\sigma(\bm{W}^\dagger\bm{x}_{i})\le 0$ for all $i=1,\,2,\,\ldots,\,n$, thanks to linear separability of the data. This in agreement with the definition of the hinge loss function satisfies that $L_{\mathcal{S}}(\bm{W}^\dagger)=0$ in \eqref{eq:opt}, which guarantees
that the algorithm converges to a \emph{global optimum}. Two critical questions arise at this point: Q1) How can we endow a non-zero `(sub-)gradient' based search direction even at a sub-optimal critical point, while having the global minima as limiting points of the algorithm? and Q2) How is it possible to guarantee convergence? 

Question Q1) can be answered by ensuring that at least one ReLU is active at a non-optimal point. Toward this objective,
motivated by recent efforts in escaping saddle points \cite{1992noise}, \cite{1996noise}, \cite{2015noise}, 
we are prompted to add a zero-mean random noise vector $\bm{\epsilon}^t\in \mathbb{R}^k$ to $\bm{W}^t\bm{x}_{i_t}\in\mathbb{R}^k$, namely the input vector to the activity indicator function of \emph{all} ReLUs. This would replace $\mathds{1}_{\{\bm{W}^t\bm{x}_{i_t}\ge \bm{0}\}}$ in the subgradient (cf.
\eqref{eq:grad}) with $\mathds{1}_{\{\bm{W}^t\bm{x}_{i_t}+\bm{\epsilon}^t\ge \bm{0}\}}$ at every iteration. In practice, Gaussian additive noise $\bm{\epsilon}^t\sim\mathcal{N}(\bm{0},\,\gamma^2\bm{I})$ with sufficiently large variance $\gamma^2>0$ works well. 

Albeit empirically effective in training ReLU networks, SGD with such architecture-agnostic injected noise into all ReLU activity indicator functions cannot guarantee convergence in general, or convergence is difficult or even impossible to establish. We shall take a different route to bypass this hurdle here, which will lead to a simple algorithm provably {convergent} to a wanted global optimum 
in a finite number of non-zero updates. This result holds {regardless} of the data distribution, initialization, network size, or the number of hidden neurons.  
Toward, to ensure convergence of our modified SGD algorithm, we carefully design the noise injection process by maintaining at least one non-zero ReLU activity indicator variable at every non-optimal critical point.

\emph{For the picked data sample $(\bm{x}_{i_t},y_{i_t})\in\mathcal{S}$ per iteration $t\ge 0$, we inject Gaussian noise $\epsilon_{j}^t\sim\mathcal{N}(0,\,\gamma^2)$ into the $j$-th ReLU activity indicator function $\mathds{1}_{\{ \bm{w}_j^\top\bm{x}_{i_t}\ge 0\}}$ in the SGD update of \eqref{eq:grad}, {if and only if} the corresponding quantity $y_{i_t}v_j\ge 0$ holds, and we repeat this for all neurons $j=1,\,2,\,\ldots,\, k$. }

Interestingly, the noise variance $\gamma^2$, admits simple choices, so long as it is selected sufficiently large matching the size of the corresponding summands $\{| {\bm{w}_j^t}^\top\bm{x}|^2\}_{j,t}$. We will build up more intuition and highlight the basic principle behind such a noise injection design shortly in Section \ref{sec:conv}, along with our formal convergence analysis.   
For implementation purposes, we summarize the novel SGD algorithm with randomly perturbed ReLU activity indicator functions in Algorithm \ref{alg:sgdn}. As far as stopping criterion is concerned, it is safe to conclude that the algorithm has converged, if there has been no non-zero update for \emph{a succession of} say, $np$ iterations, where $p>0$ is some fixed large enough integer. This holds with high probability, which depends on $p$, and $|\mathcal{N}_v^+|$ ($|\mathcal{N}_v^-|$), where the latter denotes the number of neurons with $v_j>0$ ($v_j<0$). We have the following result, whose proof is provided in Appendix \ref{proofofprop}. 
\begin{proposition}
	\label{prop:prob}
	Let $\|\bm{w}_j^t\|_2\le w_{\max}$ for all neurons $j=1,\,2,\,\ldots,\,k$, and all iterations $t\ge 0$, and consider $i_t$ cycling deterministically through $\{1,2,\ldots,n \}$.
	If there is no non-zero update after a succession of $np$ iterations, then Algorithm \ref{alg:sgdn} converges to a global optimum of $L_{\mathcal{S}}(\bm{W})$ with probability at least  $1-\left[\Phi\!\left(w_{\max}/\gamma\right) \right]^{p\min\{|\mathcal{N}_v^+|,\,|\mathcal{N}_v^-|\}}$,
	where $\Phi(z):=\left(1/\sqrt{2\pi}\right)\int_{-\infty}^z e^{-s^2} d s$ is the cumulative density function of the standardized Gaussian distribution $\mathcal{N}(0,1)$.  
\end{proposition}


%
%
%

\begin{algorithm}[t]
	\caption{Learning two-layer ReLU networks via SGD with randomly perturbed ReLU activity indicators.}
	\label{alg:sgdn}
	\begin{algorithmic}[1]
		
		\STATE {\bfseries Input:}
		Training data $\mathcal{S}=\{(\bm{x}_i, y_i)\}_{i=1}^n$,
		second layer weight vector $\bm{v}\in\mathbb{R}^k$ with at least one positive and at least one negative entry, initialization parameter $\rho\ge 0$, 
		learning rate $\eta>0$, and noise variance $\gamma^2>0$. 
		\STATE {\bfseries Initialize} $\bm{W}^0$ with $\bm{0}$, or randomly having its rows obey $\|\bm{w}_j\|_2\le \rho$. 
		\STATE {\bfseries For} {$t=0,\,1,\,2,\,\ldots$} {\bf do} 
		\STATE {\bfseries \quad~~ Pick $i_t$} uniformly at random from, or deterministically cycle through $\{1,\,2,\,\ldots,\,n\}$. 
		\STATE {\bfseries \quad~~ Update} 
		\begin{align}\label{eq:sgdn}
			\bm{W}^{t+1}=\bm{W}^t&+\eta\,\mathds{1}_{\left\{1-y_{i_t}\bm{v}^\top\sigma(\bm{W}^t\bm{x}_{i_t})> 0\right\}}\times  y_{i_t} \text{diag}\!\left(\mathds{1}_{\left\{\bm{W}^t\bm{x}_{i_t}+\bm{\epsilon}^t\ge \bm{0}\right\}}\right) \bm{v}\bm{x}_{i_t}^\top	
		\end{align}
		where per $j$-th entry of noise $\bm{\epsilon}^t\in\mathbb{R}^k$ follows 
		$\epsilon_{j}^t\sim \mathcal{N}(0,\,\gamma^2)$, if $y_{i_t}v_j\ge 0$; and $\epsilon_{j}^t=0$, otherwise. 
		\STATE {\bfseries Output:}
		$\bm{W}^{t+1}$.
	\end{algorithmic}
\end{algorithm} 

Observe that the probability in Proposition \ref{prop:prob} can be made arbitrarily close to $1$ by taking sufficiently large $p$ and/or $\gamma$. 
Regarding our proposed approach in Algorithm \ref{alg:sgdn}, three remarks are worth making.

\begin{remark}
	\label{rmk:simple}
	With the carefully designed noise injection rule, our algorithm constitutes a {non-trivial} generalization of the Perceptron or plain-vanilla SGD algorithms to learn ReLU networks. Implementing Algorithm \ref{alg:sgdn} is as easy as plain-vanilla SGD, requiring {almost negligible} extra computation overhead. Both numerically and analytically, we will demonstrate the power of our principled noise injection into partial ReLU activity indicator functions, as well as establish the \emph{optimality}, \emph{efficiency}, and \emph{generalization} performance of Algorithm \ref{alg:sgdn} in learning two-layer (over-parameterized) ReLU networks on linearly separable data.  
\end{remark}

\begin{remark}
	\label{rmk:noise}
	It is worth remaking that the random (Gaussian) noise in our proposal is \emph{solely added to} the {ReLU activity indicator functions}, \emph{rather than} to any of the hidden neurons. This is evident from the first indicator function $\mathds{1}_{\{1-y_{i_t}\bm{v}^\top\sigma(\bm{W}^t\bm{x}_{i_t})> 0\}}$ being the (sub)derivative of a hinge loss, in Step 5 of Algorithm \ref{alg:sgdn}, which is kept as it is in the plain-vanilla SGD, namely it is \emph{not affected} by the noise. 
	Moreover, our use of random noise in this way distinguishes itself from those in the vast literature for evading saddle points (see e.g., \cite{1992noise}, \cite{1996noise}, \cite{2015noise}, \cite{2018escape}), which simply add noise to either the iterates or to the (stochastic) (sub)gradients. This distinction endows our approach with the unique capability of also escaping local minima (in addition to saddle points). To the best of our knowledge, our approach is the \emph{first} of its kind in provably yet efficiently escaping local minima under suitable conditions.   
\end{remark}

\begin{remark}
	\label{rmk:efforts}
	Compared with previous efforts in learning ReLU networks (e.g., \cite{2017alon},  \cite{2017learningrelus}, \cite{2017zhong}, \cite{2017cnn}, \cite{2018am}, \cite{2018gu}), our proposed Algorithm \ref{alg:sgdn} provably converges to a global optimum in a finite number of non-zero updates, without any assumptions on the data distribution, training/network size, or initialization. This holds even in the presence of exponentially many local minima and saddle points. To the best of our knowledge, Algorithm \ref{alg:sgdn} provides the \emph{first} solution to efficiently train such a single-hidden-layer ReLU network to global optimality with a hinge loss, so long as the training samples are linearly separable. Generalizations to other objective functions based on e.g., the $\tau$-hinge loss and the smoothed hinge loss (a.k.a. polynomial hinge loss) \cite{2018adding}, as well as to multilayer ReLU networks are possible, and they are left for future research.
\end{remark}

\subsection{Convergence analysis}
\label{sec:conv}

In this section, we analyze the convergence of Algorithm \ref{alg:sgdn} for learning single-hidden-layer ReLU networks with a hinge loss criterion on linearly separable data, namely for minimizing $L_{\mathcal{S}}(\bm{W})$ in \eqref{eq:opt}. Recall since we only train the first layer having the second layer weight vector $\bm{v}\in\mathbb{R}^k$ fixed \emph{a priori}, we can assume without further loss of generality that entries of $\bm{v}$ are all non-zero. Otherwise, one can exclude the corresponding hidden neurons from the network, yielding an equivalent reduced-size NN whose second layer weight vector has all its entries non-zero. 


Before presenting our main convergence results for Algorithm \ref{alg:sgdn}, we introduce some notation. To start, let $\mathcal{N}_y^+\subseteq \{1,2,\ldots,n\}$ ($\mathcal{N}_y^-$) be the index set of data samples $\{(\bm{x}_i,y_i)\}_{1\le i\le n}$ belonging to the `positive' (`negative') class, namely whose $y_i=+1$ ($y_i=-1$).
It is thus self-evident that $\mathcal{N}_y^+\cup\mathcal{N}_y^-=\{1,2,\ldots,n\}$ and $\mathcal{N}_v^+\cup\mathcal{N}_v^-=\{1,2,\ldots,k\}$ hold under our assumptions. 
Putting our work in context, it is useful to first formally summarize the landscape properties of the objective function $L_{\mathcal{S}}(\bm{W})=(1/n)\sum_{i=1}^n\max\{0,\,1-y_i f(\bm{x}_i;\bm{W})\}$, which can help identify the challenges in learning ReLU networks. 

\begin{proposition}
	\label{prop:prop}
	Function $L_{\mathcal{S}}(\bm{W})$ has the following properties: i) it is non-convex, and ii) for each sub-optimal local minimum (that incurs a non-zero loss), there exists (at least) a datum $(\bm{x}_i,y_i)\in\mathcal{S}$ for which all ReLUs become inactive.  
\end{proposition}

The proof of Property i) in Proposition \ref{prop:prop} can be easily adapted from that of \cite[Proposition 5.1]{2018alon}, while Property ii) is just a special case of \cite[Thm. 5]{2018multilinear} for a fixed $\bm{v}$; hence they are both omitted in this paper. 

We will provide an upper bound on the number of non-zero updates that Algorithm \ref{alg:sgdn} performs until no non-zero update occurs after within a succession of say, e.g. $np$ iterations (cf. \eqref{eq:sgdn}), where $p$ is a large enough integer. This, together with the fact that all sub-optimal critical points of $L_{\mathcal{S}}(\bm{W})$ are not limiting points of Algorithm \ref{alg:sgdn} due to the Gaussian noise injection with a large enough variance $\gamma^2>0$ at every iteration, will guarantee convergence of Algorithm \ref{alg:sgdn} to a global optimum of $L_\mathcal{S}(\bm{W})$.
Specifically, the main result is summarized in the following theorem.

\begin{theorem}[\textbf{Optimality}]
	\label{th:main}
	If all rows of the initialization $\bm{W}^0$ satisfy $\|\bm{w}_j^0\|_2\le \rho$ for any constant $\rho\ge 0$, and the second layer weight vector $\bm{v}\in\mathbb{R}^k$ is kept fixed with both positive and negative (but non-zero) entries,
	then 
	Algorithm \ref{alg:sgdn} with some constant step size $\eta>0$ converges to a global minimum of $L_{\mathcal{S}}(\bm{W})$ after performing at most $T_k$ non-zero updates, where for $v_{\min}=\min_{1\le j\le k}|v_j|$ it holds that
	\begin{align}
		T_k:=&~
		\frac{k}{\eta v_{\min}^2}\bigg[\left(\eta\left\|\bm{v}\right\|_2^2+2\right)\left\|\bm{\omega}^\ast\right\|_2^2+2\rho v_{\min}\left\|\bm{v}\right\|_2^2
		+\sqrt{2 \rho v_{\min} \left\|\bm{\omega}^\ast\right\|_2\left( \eta\left\|\bm{v}\right\|_2^2+2\right)}\left\|\bm{\omega}^\ast\right\|_2\bigg].
	\end{align}
	In particular, if $\bm{W}^0=\bm{0}$, then Algorithm \ref{alg:sgdn} converges to a global optimum after at most $T_k^0:=\frac{k}{\eta v_{\min}^2}\big (\eta \|\bm{v}\|_2^2+2\big)\|\bm{\omega}^\ast\|_2^2$ non-zero updates. 
\end{theorem}

Regarding Theorem \ref{th:main}, a couple of observations are of interest. The developed Algorithm \ref{alg:sgdn} converges to a globally optimal solution of the non-convex optimization \eqref{eq:opt} within a finite number of non-zero updates, which implicitly corroborates the ability of Algorithm \ref{alg:sgdn} to escape sub-optimal local minima, as well as saddle points. 
This holds regardless of the underlying data distribution $\mathcal{D}$, the number $n$ of training samples, the number $k$ of hidden neurons, or even the initialization $\bm{W}^0$. It is also worth highlighting that the number $T_k$ of non-zero updates does not depend on the dimension $d$ of input vectors, but it scales with $k$ (in the worst case), and it is inversely proportional to the step size $\eta>0$. Recall that
the worst-case bound for SGD learning of leaky-ReLU networks with initialization $\bm{W}^0=\bm{0}$ is \cite[Thm. 2]{2018alon} 
\begin{equation}
	T_{\rm leaky}^\alpha\le \frac{\|\bm{\omega}^\ast\|_2^2}{\alpha^2}\left(1+ \frac{1	}{\eta \|\bm{v}\|_2^2}\right)\label{eq:lrelu}
\end{equation}
where again, $\bm{\omega}^\ast$ denotes an optimal linear classifier obeying $\mathbb{P}_{(\bm{x},y)\sim \mathcal{D}}(y\, {\bm{\omega}^\ast}^\top \bm{x} \ge 1)=1$. Clearly, the upper bound above \emph{does not} depend on $k$. This is due to the fact that the loss function corresponding to learning leaky-ReLU networks has \emph{no} bad local minima, since all critical points are global minima. This is in sharp contrast with the loss function associated with learning ReLU networks investigated here, which generally involves \emph{infinitely many} bad local minima! On the other hand, the bound in \eqref{eq:lrelu} scales inversely proportional with the quadratic `leaky factor' $\alpha^2$ of leaky ReLUs. 
This motivates having $\alpha\to 0$, which corresponds to letting the leaky ReLU approach the ReLU. In such a case, \eqref{eq:lrelu} would yield a worst-case bound of infinity for learning ReLU networks, corroborating the challenge and impossibility of learning ReLU networks by `plain-vanilla' SGD. Indeed, the gap between $T_k^0$ in Theorem \ref{th:main} and the bound in \eqref{eq:lrelu} is the price for being able to escape local minima and saddle points paid by our noise-injected SGD Algorithm \ref{alg:sgdn}. 
Last but not least, Theorem \ref{th:main} also suggests that for a given network and a fixed step size $\eta$, Algorithm \ref{alg:sgdn} with $\bm{W}^0=\bm{0}$ works well too.   

We briefly present the main ideas behind the proof of Theorem \ref{th:main} next, but delegate the technical details to Appendix \ref{proofoftheorem}. Our proof mainly builds upon the convergence proof of the classical Perceptron algorithm (see e.g., \cite[Thm. 9.1]{book2014shai}), and it is also inspired by that of \cite[Thm. 1]{2018alon}. Nonetheless, the novel approach of performing SGD with principled noise injection into the ReLU activity indicator functions distinguishes itself from previous efforts. Since we are mainly interested in the (maximum) number of non-zero updates to be performed until convergence, we will assume for notational convenience that all iterations $t\ge 0$ in \eqref{eq:sgdn} of Algorithm \ref{alg:sgdn} perform a non-zero update. This assumption is made without loss of generality. To see this, since after the algorithm converges, one can always re-count the number of effective iterations that correspond to a non-zero update and re-number them by $t=0,1,\ldots.$

Our main idea is to demonstrate that \emph{every single} non-zero update of the form \eqref{eq:sgdn} in Algorithm \ref{alg:sgdn} makes a \emph{non-negligible} progress in bringing the current iterate $\bm{W}^t\in\mathbb{R}^{k\times d}$ toward some global optimum $\bm{\Omega}^\ast\in\mathbb{R}^{k\times d}$ of $L_{\mathcal{S}}(\bm{W})$, constructed based on the linear classifier weight vector $\bm{\omega}^\ast$. Specifically, as in the convergence proof of the Perceptron algorithm, we will establish separately a lower bound on the term $\langle \bm{W}^t,\bm{\Omega}^\ast\rangle_F $, which is the so-termed Frobenius inner product, performing a component-wise inner product of two same-size matrices as though they are vectors; and, an upper bound on the norms $\|\bm{W}_t\|_F$ and $\|\bm{\Omega}^\ast\|_F$. Both bounds will be carefully expressed as functions of the number $t$ of \emph{performed} non-zero updates. Recalling the Cauchy-Schwartz inequality 
$|\langle \bm{W}^t,\bm{\Omega}^\ast\rangle_F |\le \|\bm{W}_t\|_F \|\bm{\Omega}^\ast\|_F$, the lower bound of $|\langle \bm{W}^t,\bm{\Omega}^\ast\rangle_F |$ cannot grow larger than the upper bound on $\|\bm{W}_t\|_F \|\bm{\Omega}^\ast\|_F$. Since every non-zero update brings the lower and upper bounds closer by a non-negligible amount, the worst case (in terms of the number of non-zero updates) is to have the two bounds equal at convergence, i.e., $|\langle \bm{W}^t,\bm{\Omega}^\ast\rangle_F |=\|\bm{W}_t\|_F \|\bm{\Omega}^\ast\|_F$. To arrive at this equality, we are able to deduce an upper bound (due to a series of inequalities used in the proof to produce a relatively clean bound) on the number of non-zero updates by solving a univariate quadratic equality. 

It will become clear in the proof that injecting random noise into just a subset of (rather than all) ReLU activity indicator functions enables us to leverage two key inequalities, namely, $\sum_{j=1}^k y_{i_t}v_j \sigma({ \bm{w}_j^t}^\top\bm{x}_{i_t}) <1$ and $y_i{ \bm{\omega}^\ast}^\top\bm{x}_i \ge 1$ for all data samples $(\bm{x}_i,y_i)\in\mathcal{S}$. These inequalities uniquely correspond to whether an update is non-zero or not. In turn, this characterization is indeed the key to establishing the desired lower and upper bounds for the two quantities on the two sides of the Cauchy-Schwartz inequality, a critical ingredient of our convergence analysis.      




\subsection{Lower bound}

Besides the worst-case upper bound given in Theorem \ref{th:main}, we also provide a lower bound on the number of non-zero updates required by Algorithm \ref{alg:sgdn} for convergence, which is summarized in the following theorem. The proof is provided in Appendix \ref{proofoflower}.  
\begin{theorem}[\textbf{Lower bound}]
	\label{th:lower}
	Under the conditions of Theorem \ref{th:main}, consider Algorithm \ref{alg:sgdn} with initialization $\bm{W}^0=\bm{0}$. Then for any $d>0$, there exists a set of linearly separable data samples on which Algorithm \ref{alg:sgdn} performs at least ${\|\bm{\omega}^\ast\|_2^2}\big/\!\left({\eta \|\bm{v}\|_2^2}\right)$ non-zero updates to optimally train a single-hidden-layer ReLU network. 
\end{theorem}

The lower bound on the number of non-zero updates to be performed in Theorem \ref{th:lower} matches that for learning single-hidden-layer leaky-ReLU networks initialized from zero \cite[Thm. 4]{2018alon}. On the other hand, it is also clear that the worst-case bound established in Theorem \ref{th:main} is (significantly) loose than the lower bound here. The gap between the two bounds (in learning ReLU versus leaky ReLU networks) is indeed the price we pay for escaping bad local minima and saddle points through our noise-injected SGD approach.

\begin{figure*}[t]
	\centering
	\begin{subfigure}[t]{1\textwidth}
		\includegraphics[height=.2\textheight,width=\textwidth]{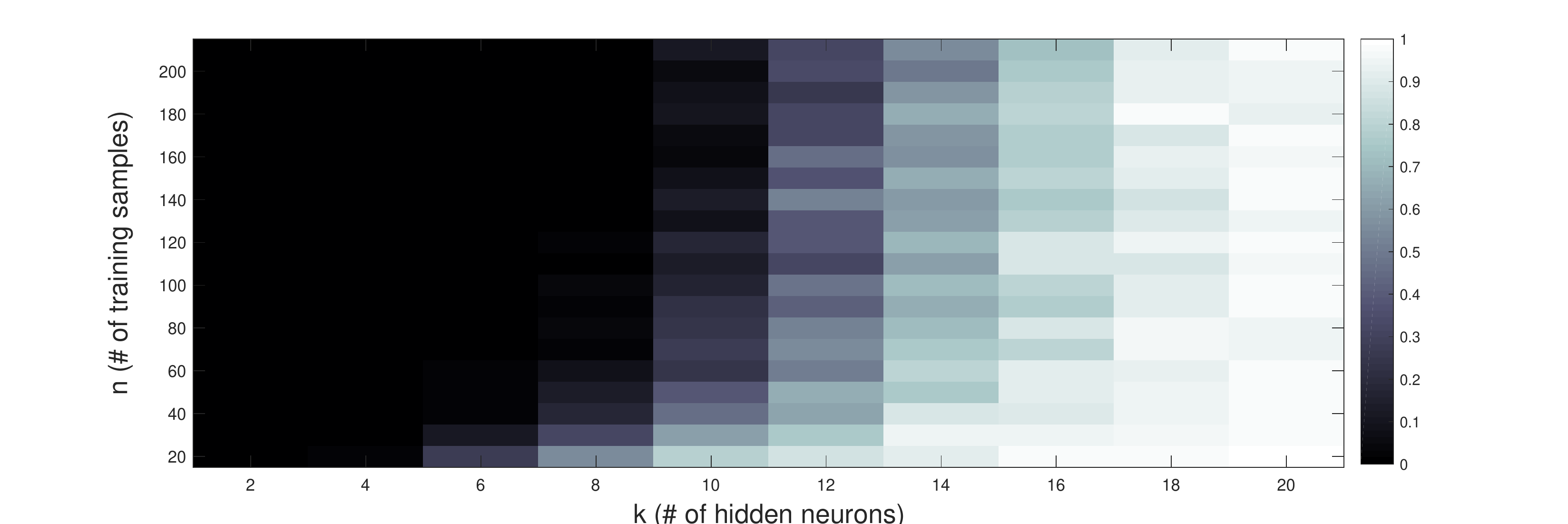} 
	\end{subfigure}
	\begin{subfigure}[t]{1\textwidth}
		\includegraphics[height=.2\textheight,width=\textwidth]{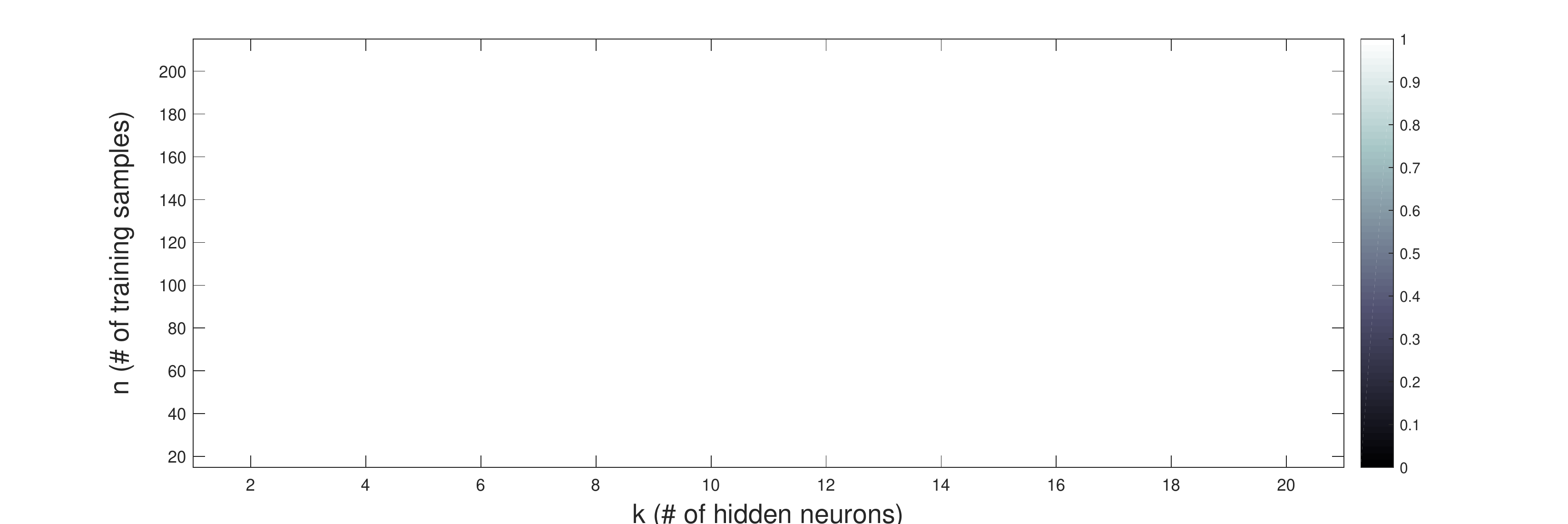}
	\end{subfigure}
	\caption{Empirical success rates of plain-vanilla SGD (top panel) and the Algorithm \ref{alg:sgdn} (bottom panel) for learning two-layer ReLU networks of $k$ hidden units on $n$ randomly generated data samples of dimension $d=128$.
	}
	\label{fig:srate}
\end{figure*}

\section{Generalization}
\label{sec:gene}

In this section, we investigate the generalization performance of training (possibly over-parameterized) ReLU networks using Algorithm \ref{alg:sgdn} with randomly perturbed ReLU activity indicator functions. Toward this objective, we will rely on compression generalization bounds, specifically for the $0/1$ classification error as in \eqref{eq:error}  \cite{1986compression}.

Recall that our ReLU network has $k$ hidden units, and a fixed second-layer weight $\bm{v}\in\mathbb{R}^k$. Stressing the number of ReLUs in the subscript, let
$G_k^{\rm Alg1}(\mathcal{S};\bm{W}^0)$ denote the classifier 
obtained by training the network over training set $\mathcal{S}$ using Algorithm \ref{alg:sgdn} 
with initialization $\bm{W}^0$ having rows obeying $\{\|\bm{w}_j^0\|_2\le \rho\}_{j=1}^k$. Let also $\mathcal{G}_k$ denote the set of all classifiers $\{G_k\}$ obtained using any $\mathcal{S}$ and any $\bm{W}^0$, not necessarily those employed by Algorithm \ref{alg:sgdn}. 

Suppose now that Algorithm \ref{alg:sgdn} has converged after $\tau_k\le T_k$ non-zero updates, as per Theorem \ref{th:main}. And let $(\bm{x}_{i_1},\,\bm{x}_{i_2},\,\ldots,\,\bm{x}_{i_{\tau_k}} )$ be the $\tau_k$-tuple of training data from $\mathcal{S}$ randomly picked by SGD  iterations of Algorithm \ref{alg:sgdn}. To exemplify the $\tau_k$-tuple used per realization of Algorithm \ref{alg:sgdn}, we write $G_k^{\rm Alg1}(\mathcal{S};\bm{W}^0)=\Gamma_{\bm{W}^0}( \bm{x}_{i_1},\,\bm{x}_{i_2},\,\ldots,\,\bm{x}_{i_{\tau_k}})$. Since $\tau_k$ can be smaller than $n$, function $\Gamma_{\bm{W}^0}$ and thus $G_k^{\rm Alg1}(\mathcal{S};\bm{W}^0)$ rely on compressed (down to size $\tau_k$) versions of the $n$-tuples comprising the set $\mathcal{H}_k$
\cite[Definition 30.4]{book2014shai}. Let $\mathcal{S}_{\tau_k}^c:=\{i|i\in\{1,\,2,\,\ldots,\,n   \} \backslash \{i_1,\,i_2,\,\ldots,\,i_{\tau_k} \}\} $ be the subset of training data not picked by SGD to yield $G_k^{\rm Alg1}(\mathcal{S};\bm{W}^0)$; and correspondingly, let $R_{\mathcal{D}}(G_{k}^{\rm Alg1}(\mathcal{S};\bm{W}^0))$ denote the ensemble risk associated with $G_{k}^{\rm Alg1}$, and $\hat{R}_{\mathcal{S}_{\tau_k}^c}(G_{k}^{\rm Alg1}(\mathcal{S};\bm{W}^0))$   the empirical risk associated with the complement training set, namely $\mathcal{S}_{\tau_k}^c$. With these notational conventions, our next result follows from \cite[Thm. 30.2]{book2014shai}.

\begin{theorem}[\textbf{Compression bound}]
	\label{th:compression}
	If $n\ge 2\tau_k$, then the following inequality holds with probability of at least $1-\delta$ over the choice of $\mathcal{S}$ and $\bm{W}^0$
	\begin{align}
		\mathcal{R}_{\mathcal{D}}(G_{k}^{\rm Alg1}(\mathcal{S};\bm{W}^0))\le \hat{R}_{\mathcal{S}_{\tau_k}^c}(G_{k}^{\rm Alg1}(\mathcal{S};\bm{W}^0))+\sqrt{\hat{R}_{\mathcal{S}_{\tau_k}^c}(G_{k}^{\rm Alg1}(\mathcal{S};\bm{W}^0)) \frac{4 \tau_k \log (n/\delta)}{n}}+\frac{8\tau_k\log(n/\delta)}{n }.\label{eq:compression}
	\end{align} 
\end{theorem}

Regarding Theorem \ref{th:compression}, two observations are in order. The bound in \eqref{eq:compression} is non-asymptotic but as $n\to\infty$, the last two terms on the right-hand-side vanish, implying that
the ensemble risk $R_{\mathcal{D}}(G_{k}^{\rm Alg1}(\mathcal{S};\bm{W}^0))$ is upper bounded by the empirical risk $\hat{R}_{\mathcal{S}_{\tau_k}^c}(G_{k}^{\rm Alg1}(\mathcal{S};\bm{W}^0))$. Moreover, once the SGD iterations in Algorithm \ref{alg:sgdn} converge, we can find the complement training set $\mathcal{S}_{\tau_k}^c$, and thus $\hat{R}_{\mathcal{S}_{\tau_k}^c}(G_{k}(\mathcal{S};\bm{W}^0))$ can be determined.
After recalling that $\hat{R}_{\mathcal{S}_{\tau_k}^c}(G_k^{\rm Alg1}(\mathcal{S};\bm{W}^0))=0$ holds at a global optimum of $L_{\mathcal{S}}$ by Theorem \ref{th:main}, we obtain from Theorems \ref{th:main} and \ref{th:compression} the following corollary. 
\begin{corollary}\label{co1}
	If $n\ge 2\tau_k$, and all rows of the initialization satisfy $\{\|\bm{w}_j^0\|_2\le \rho\}_{j=1}^k$, then the following holds with probability at least $1-\delta$ over the choice of $\mathcal{S}$
	\begin{equation}\label{eq:corollary}
		R_{\mathcal{D}}(G_k^{\rm Alg1}(\mathcal{S};\bm{W}^0))\le \frac{8T_k\log(n/\delta) }{n}
	\end{equation}
	where $T_k$ is given in Theorem \ref{th:main}. 
\end{corollary}

Expressed differently, the bound in \eqref{eq:corollary} suggests that in order to guarantee a low generalization error, one requires \emph{in the worst case} about $n=\mathcal{O}(k^2\|\bm{\omega}^\ast\|_2^2)$ training data to \emph{reliably}
learn a two-layer ReLU network of $k$ hidden neurons. This holds true \emph{despite} the fact that Algorithm \ref{alg:sgdn} can achieve a zero training loss \emph{regardless of} the training size $n$. One implication of Corollary \ref{co1} is a fundamental difference in the sample complexity for generalization between training a ReLU network (at least in the worst case), versus training a $\alpha$-leaky ReLU network ($0<\alpha<1$), which at most needs $n=\mathcal{O}(\|\bm{\omega}^\ast\|_2^2/\alpha^2)$ data to be trained via SGD-type algorithms.

\section{Numerical Tests}
\label{sec:test}

To validate our theoretical results, this section evaluates the empirical performance of Algorithm \ref{alg:sgdn} using both synthetic data and real data. 
To benchmark Algorithm \ref{alg:sgdn}, we also simulated the plain-vanilla SGD. To compare between the two algorithms as fair as possible, the same initialization $\bm{W}^0$, constant step size $\eta>0$, and data random sampling scheme were employed. 
For reproducibility, the Matlab code of Algorithm \ref{alg:sgdn} is publicly available at \url{https://gangwg.github.io/RELUS/}.

\subsection{Synthetic data}

We consider first two synthetic tests using data generated from Gaussian as well as uniform distributions. In the first test, feature vectors  $\{\bm{x}_i\in\mathbb{R}^d \}_{i=1}^n$ were sampled i.i.d. from a standardized Gaussian distribution $\mathcal{N}(\bm{0},\bm{I}_d)$, and classifier $\bm{\omega}^\ast\in\mathbb{R}^d$ was drawn from $\mathcal{N}(\bm{0},\bm{I}_d)$. Labels $y_i\in \{+1,-1 \}$ were generated according to $y_i={\rm sgn}({\bm{\omega}^\ast}^\top\bm{x}_i)$. To further yield $y_i{\bm{\omega}^\ast}^\top\bm{x}_i\ge 1$ for all $i=1,\,2,\,\ldots,\,n$, we normalized $\bm{\omega}^\ast$ by the smallest number among  
$\{y_i{\bm{\omega}^\ast}^\top\bm{x}_i\}_{i=1}^n$. 
We performed $100$ independent experiments with $d=128$, and over a varying set of $n\in \{20,40,\ldots,200 \}$ training samples using ReLU networks comprising $k\in \{2,4,\ldots,20 \}$ hidden neurons.  
The second layer weight vector $\bm{v}\in\mathbb{R}^k$ was kept fixed with the first $\lfloor k/2\rfloor$ entries being $+1$ and the remaining being $-1$.  
For fixed $n$ and $k$, each experiment used a random initialization generated from $\mathcal{N}(\bm{0},0.01\bm{I})$, step size $\eta=0.01$, and noise variance $\gamma=100$, along with a maximum of $5,000$ effective data passes.

Figure \ref{fig:srate} depicts our results, where we display success rates of the plain-vanilla SGD (top panel) and our noise-injected SGD in Algorithm \ref{alg:sgdn} (bottom panel); each plot presents results obtained from the $100$ experiments. Within each plot, a white square signifies that $100\%$ of the trials were successful, meaning that the learned ReLU network yields a training loss $L_{\{(\bm{x}_i,y_i) \}_{i=1}^n}(\bm{W}^T)\le 10^{-10}$, while black squares indicate $0\%$ success rates. It is evident that the developed Algorithm \ref{alg:sgdn} trained all considered ReLU networks to global optimality, while plain-vanilla SGD can get stuck with bad local minima, for small $k$ in particular. The bottom panel confirms that Algorithm \ref{alg:sgdn} achieves optimal learning of single-hidden-layer ReLU networks on separable data, regardless of the network size, the number of training samples, and the initialization. The top panel however, suggests that learning ReLU networks becomes easier with plain-vanilla SGD as $k$ grows larger, namely as the network becomes `more over-parameterized.'    

\begin{figure*}
	\centering
	\begin{subfigure}[t]{0.48\textwidth}
		\includegraphics[width=\textwidth]{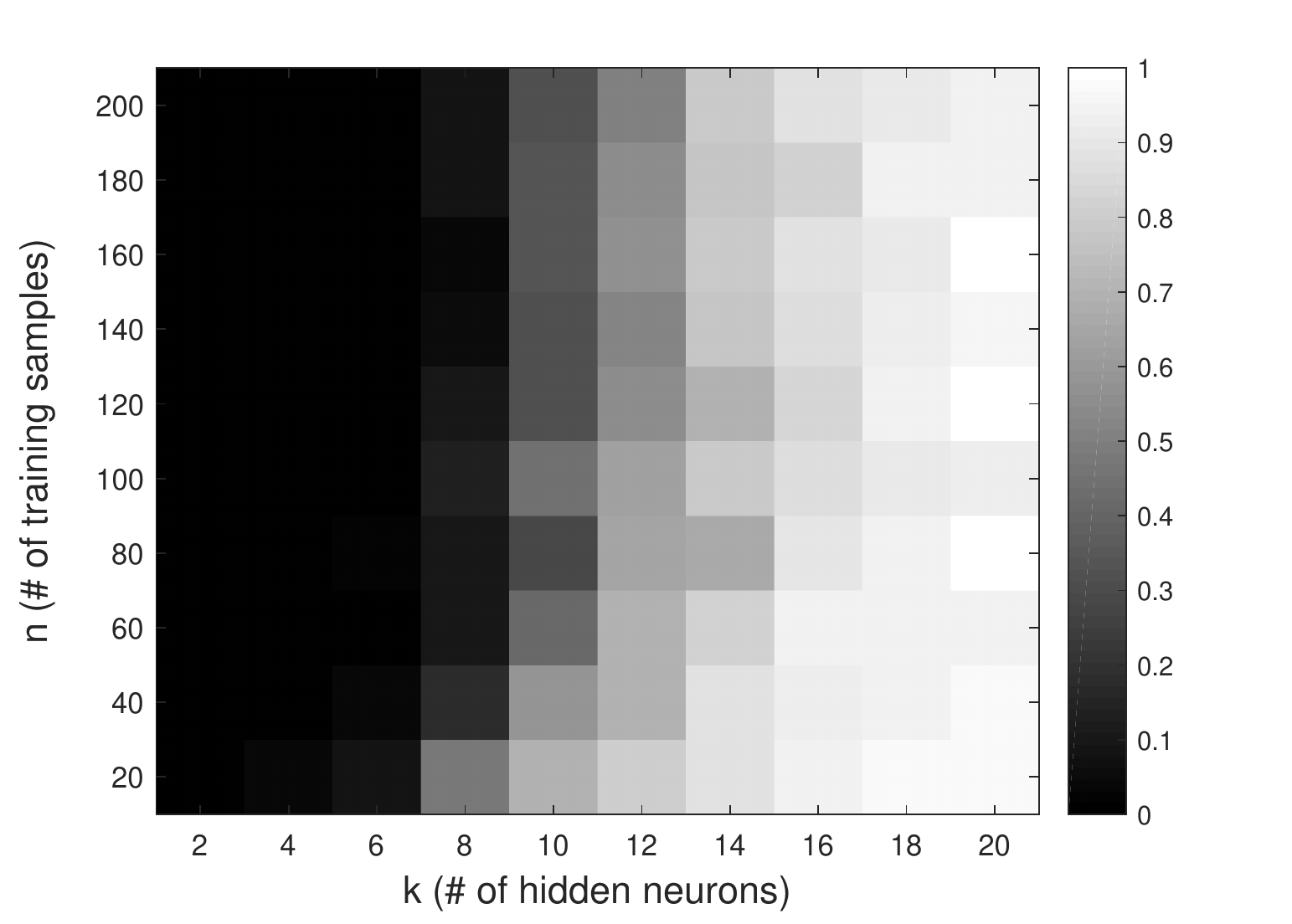}  
	\end{subfigure}
	\begin{subfigure}[t]{0.48\textwidth}
		\includegraphics[width=\textwidth]{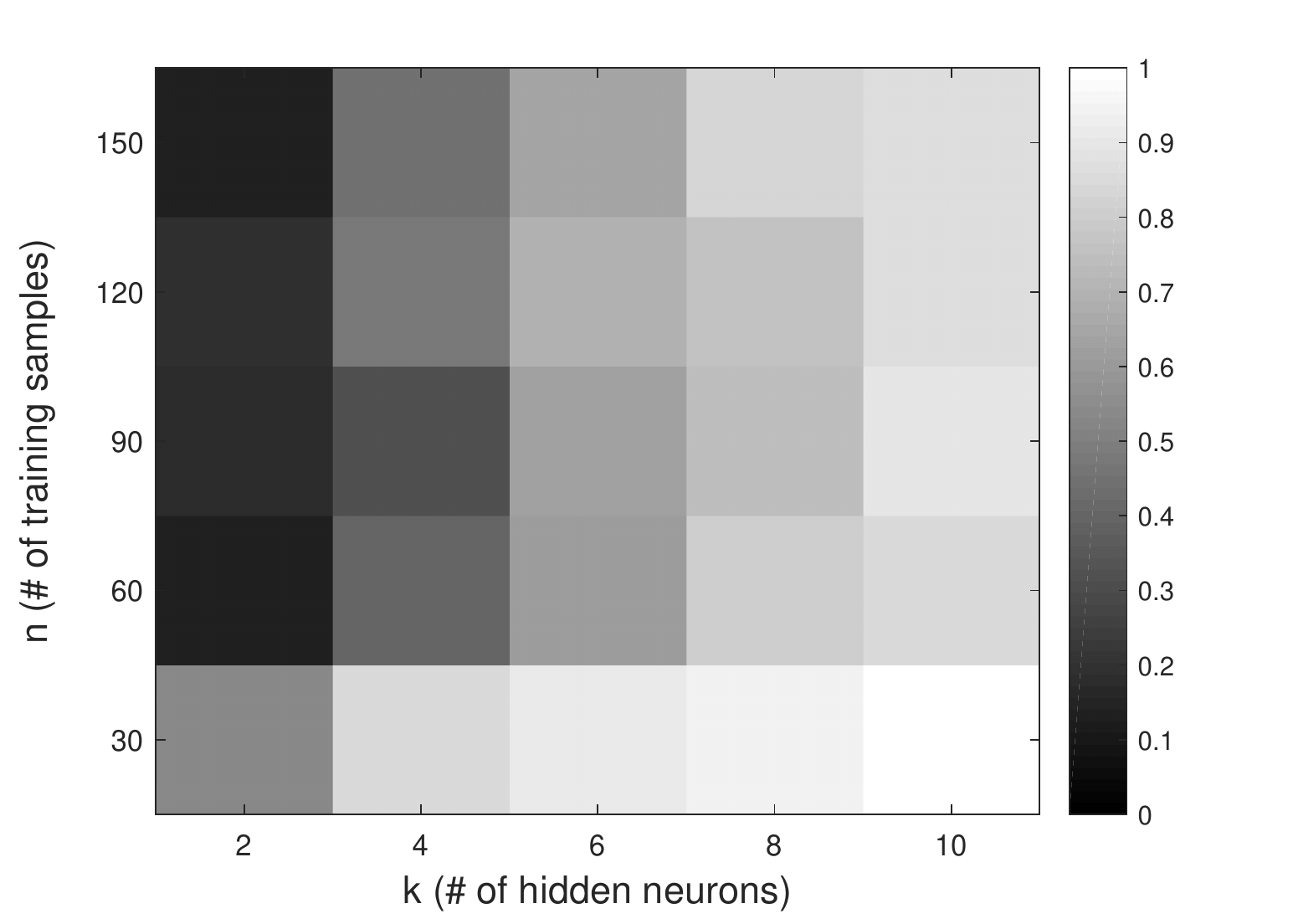}     
	\end{subfigure}
	\caption{Empirical success rates of the `plain-vanilla' SGD for learning two-layer ReLU networks of $k$ hidden units on $n$ data samples of dimension: (left) $d=128$ generated from uniform distribution $\mathcal{U}[-\bm{1}_{128},\bm{1}_{128}]$, and (right) $d=4$ from the UCI machine learning repository \cite{uci}.
	}
	\label{fig:added}
\end{figure*}

We repeated the first test using synthetic data $\{\bm{x}\in\mathbb{R}^{128} \}_{i=1}^n$ as well as classifier $\bm{\omega}\in\mathbb{R}^{128}$ generated i.i.d. from the uniform distribution $\mathcal{U}[-\bm{1}_{128},\,\bm{1}_{128}]$. All other settings were kept the same. Success rates of plain-vanilla SGD are plotted in Fig. \ref{fig:added} (left panel), while those of the proposed Algorithm \ref{alg:sgdn} are omitted, as they are $100\%$ successful in all simulated tests.

\subsection{Real data}

Performance of Algorithm \ref{alg:sgdn} for training (over-)parameterized ReLU networks is further corroborated using two real datasets: \emph{iris} in UCI's machine learning repository \cite{uci}, and \emph{MNIST} images 
\footnote{Downloaded from http://yann.lecun.com/exdb/mnist/.
}. The iris dataset contains $150$ four-dimensional feature vectors belonging to three classes. To obtain a two-class linearly separable dataset, the first-class data vectors were relabeled $+1$, while the remaining were relabeled $-1$. We performed $100$ independent experiments over a varying set of $n\in \{30,\,60,\,90,\,120,\,150 \}$ training samples using ReLU networks with $k\in\{2,\,4,\,6,\,8,\,10 \}$ hidden neurons. Gaussian initialization from $\mathcal{N}(\bm{0},\bm{I})$, step size $\eta=0.1$, noise variance $\gamma=10$, and a maximum of $100$ effective data passes were simulated. Success rates of plain-vanilla SGD are given in Fig. \ref{fig:added} (right). Again, Algorithm \ref{alg:sgdn} achieves a $100\%$ success rate in all simulated settings.

\begin{figure*}[t]
	\centering
	\begin{subfigure}[t]{1\textwidth}
		\includegraphics[height=.18\textheight,width=\textwidth]{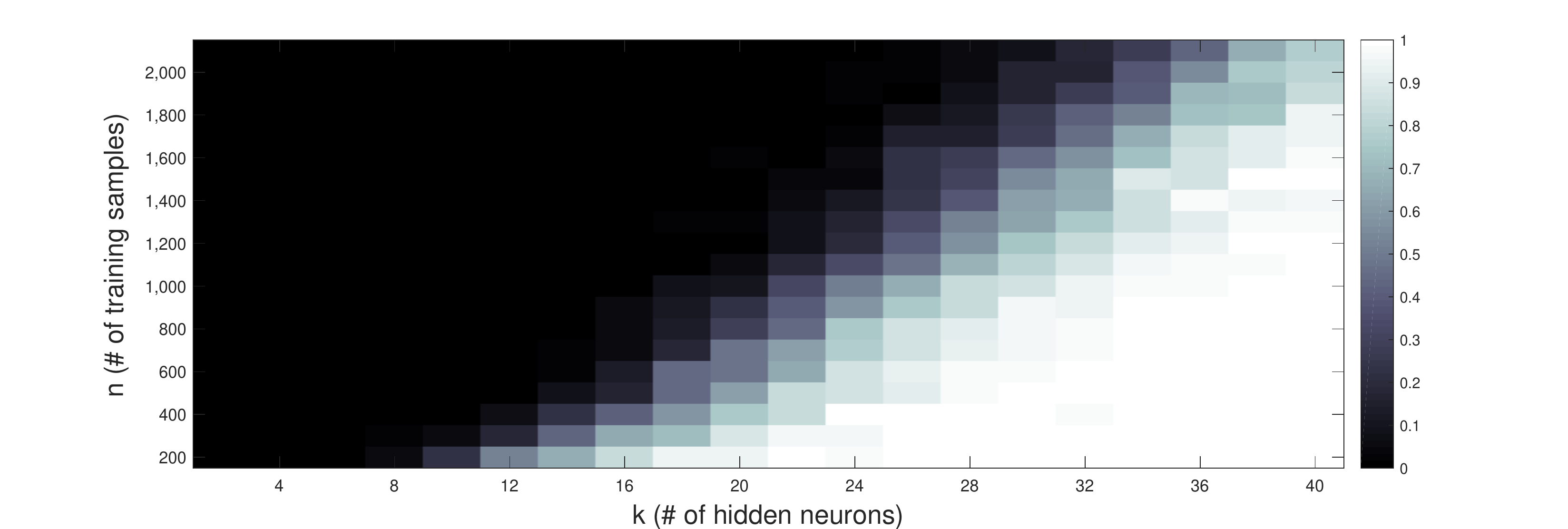} 
	\end{subfigure}
	\begin{subfigure}[t]{1\textwidth}
		\includegraphics[height=.18\textheight,width=\textwidth]{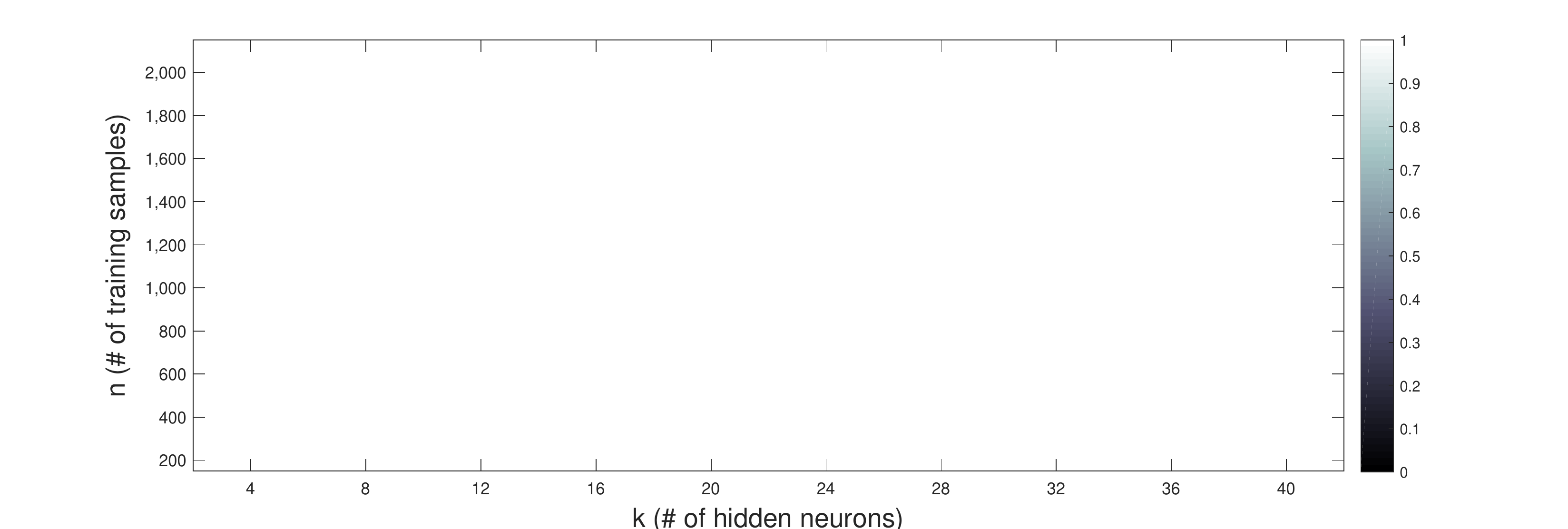}
	\end{subfigure}
	\caption{Empirical success rates of plain-vanilla SGD (top panel) and Algorithm \ref{alg:sgdn} (bottom panel) for learning two-layer ReLU networks of $k$ hidden units on $n$ MNIST images of digits $3$ and $5$.
	}
	\label{fig:mnist}
\end{figure*}

The linearly separable MNIST dataset collects $2,000$ images of digits $3$ (labeled $+1$) and $5$ (labeled $-1$),
each having dimension $784$. We performed $100$ independent experiments over a varying set of $n\in \{200,\,400,\,\ldots,\,2,000 \}$ training samples using ReLU networks with $k\in\{2,\,4,\,\ldots,\,40 \}$ hidden neurons. The constant step size of both plain-vanilla SGD and Algorithm \ref{alg:sgdn} was set to $\eta=0.001$ ($\eta=0.01$) when the ReLU networks have $k\le 4$ ($k>4$) hidden units, while
the noise variance in Algorithm \ref{alg:sgdn} was set to $\gamma=10$. Similar to the first experiment on randomly generated data, we plot success rates of the plain-vanilla SGD (top panel) and our noise-injected SGD (bottom panel) algorithms over training sets of MNIST images in 
Figure \ref{fig:mnist}. It is self-evident that Algorithm \ref{alg:sgdn} achieved a $100\%$ success rate under all testing conditions, which confirms our theoretical results in Theorem \ref{th:main}, and it 
markedly improves upon its plain-vanilla SGD alternative.

\section{Conclusions}
\label{sec:conc}

This paper approached the task of training ReLU networks from a non-convex optimization point of view. Focusing on the task of binary classification with a hinge loss criterion, this contribution put forth the first algorithm that can provably yet efficiently train any single-hidden-layer ReLU network to global optimality, provided that the data are linearly separable. The algorithm is as simple as plain-vanilla SGD, but it is able to exploit the power of random additive noise to break `optimality' of the SGD learning process at any sub-optimal critical point. We established an upper and a lower bound on the number of non-zero updates that the novel algorithm requires for convergence to a global optimum. Our result holds regardless of the underlying data distribution, network/training size, or initialization. We further developed generalization error bounds for two-layer NN classifiers with ReLU activations, which provide the first theoretical guarantee for the generalization behavior of ReLU networks trained with SGD. 
A comparison of such bounds with those of a leaky ReLU network reveals a key difference between optimally learning a ReLU network versus that of a leaky ReLU network in the sample complexity required for generalization.

Since analysis, comparisons, and corroborating tests focus on single-hidden-layer networks with a hinge loss criterion here, our future work will naturally aim at generalizing the novel noise-injection design to SGD for multilayer ReLU networks, and considering alternative loss functions, and generalizations to (multi-)kernel based approaches.    

\bibliographystyle{IEEEtranS}
\bibliography{dl}

\begin{thebibliography}{10}
\providecommand{\url}[1]{#1}
\csname url@samestyle\endcsname
\providecommand{\newblock}{\relax}
\providecommand{\bibinfo}[2]{#2}
\providecommand{\BIBentrySTDinterwordspacing}{\spaceskip=0pt\relax}
\providecommand{\BIBentryALTinterwordstretchfactor}{4}
\providecommand{\BIBentryALTinterwordspacing}{\spaceskip=\fontdimen2\font plus
\BIBentryALTinterwordstretchfactor\fontdimen3\font minus
  \fontdimen4\font\relax}
\providecommand{\BIBforeignlanguage}[2]{{%
\expandafter\ifx\csname l@#1\endcsname\relax
\typeout{** WARNING: IEEEtranS.bst: No hyphenation pattern has been}%
\typeout{** loaded for the language `#1'. Using the pattern for}%
\typeout{** the default language instead.}%
\else
\language=\csname l@#1\endcsname
\fi
#2}}
\providecommand{\BIBdecl}{\relax}
\BIBdecl

\bibitem{1996noise}
G.~An, ``The effects of adding noise during backpropagation training on a
  generalization performance,'' \emph{Neural Comput.}, vol.~8, no.~3, pp.
  643--674, Apr. 1996.

\bibitem{1996exponential}
P.~Auer, M.~Herbster, and M.~K. Warmuth, ``Exponentially many local minima for
  single neurons,'' in \emph{Adv. in Neural Inf. Process. Syst.}, Denver,
  Colorado, Nov. 27--Dec. 2, 1995, pp. 316--322.

\bibitem{1989nphard}
A.~Blum and R.~L. Rivest, ``Training a 3-node neural network is
  {NP}-complete,'' in \emph{Adv. in Neural Inf. Process. Syst.}, Cambridge,
  Massachusetts, Aug. 3--5, 1988, pp. 494--501.

\bibitem{2017alon}
A.~Brutzkus and A.~Globerson, ``Globally optimal gradient descent for a
  {C}onv{N}et with {G}aussian inputs,'' in \emph{Intl. Conf. on Mach. Learn.},
  vol.~70, Sydney, Australia, Aug. 6--11, 2017.

\bibitem{2018alon}
A.~Brutzkus, A.~Globerson, E.~Malach, and S.~Shalev-Shwartz, ``{SGD} learns
  over-parameterized networks that provably generalize on linearly separable
  data,'' in \emph{Intl. Conf. on Learn. Rep.}, Vancouver, BC, Canada, Apr.
  30--May 3, 2018.

\bibitem{1990clarke}
F.~H. Clarke, \emph{Optimization and {N}onsmooth {A}nalysis}.\hskip 1em plus
  0.5em minus 0.4em\relax SIAM, 1990, vol.~5.

\bibitem{uci}
\BIBentryALTinterwordspacing
D.~Dheeru and E.~Karra~Taniskidou, ``{UCI} machine learning repository,'' 2017.
  [Online]. Available: \url{http://archive.ics.uci.edu/ml.}
\BIBentrySTDinterwordspacing

\bibitem{2017cnn}
S.~S. Du, J.~D. Lee, Y.~Tian, B.~Poczos, and A.~Singh, ``Gradient descent
  learns one-hidden-layer {CNN}: {D}on't be afraid of spurious local minima,''
  in \emph{Intl. Conf. on Mach. Learn.}, vol.~80, Stockholm, Sweden, July
  10--15, 2018, pp. 1338--1347.

\bibitem{2017improved}
G.~K. Dziugaite and D.~M. Roy, ``Computing nonvacuous generalization bounds for
  deep (stochastic) neural networks with many more parameters than training
  data,'' \emph{arXiv:1703.11008}, 2017.

\bibitem{2018chi}
H.~Fu, Y.~Chi, and Y.~Liang, ``Guaranteed recovery of one-hidden-layer neural
  networks via cross entropy,'' \emph{arXiv:1802.06463}, 2018.

\bibitem{2015noise}
R.~Ge, F.~Huang, C.~Jin, and Y.~Yuan, ``Escaping from saddle points ---
  {O}nline stochastic gradient for tensor decomposition,'' in \emph{Conf. on
  Learn. Theory}, vol.~40, Paris, France, July 3--6, 2015, pp. 797--842.

\bibitem{2010understanding}
X.~Glorot and Y.~Bengio, ``Understanding the difficulty of training deep
  feedforward neural networks,'' in \emph{Intl. Conf. on Artif. Intell. and
  Stat.}, Sardinia, Italy, May 13--15, 2010, pp. 249--256.

\bibitem{dlbook}
I.~Goodfellow, Y.~Bengio, and A.~Courville, \emph{Deep {L}earning}.\hskip 1em
  plus 0.5em minus 0.4em\relax Cambridge: MIT press, 2016, vol.~1.

\bibitem{1992localminima}
M.~Gori and A.~Tesi, ``On the problem of local minima in backpropagation,''
  \emph{IEEE Trans. Pattern Anal. Mach. Intell.}, no.~1, pp. 76--86, Jan. 1992.

\bibitem{1992noise}
L.~Holmstrom and P.~Koistinen, ``Using additive noise in back-propagation
  training,'' \emph{IEEE Trans. Neural Netw.}, vol.~3, no.~1, pp. 24--38, Jan.
  1992.

\bibitem{2018am}
G.~Jagatap and C.~Hedge, ``Learning {R}e{LU} networks via alternating
  minimization,'' \emph{arXiv:1806.07863}, 2018.

\bibitem{kalan2019fitting}
S.~M.~M. Kalan, M.~Soltanolkotabi, and A.~S. Avestimehr, ``Fitting {ReLUs via
  SGD and Quantized SGD},'' \emph{arXiv:1901.06587}, 2019.

\bibitem{2016nolocalminima}
K.~Kawaguchi, ``Deep learning without poor local minima,'' in \emph{Adv. in
  Neural Inf. Process. Syst.}, Barcelona, Spain, Dec. 5--10, 2016, pp.
  586--594.

\bibitem{2019elimination}
K.~Kawaguchi and L.~P. Kaelbling, ``Elimination of all bad local minima in deep
  learning,'' \emph{arXiv:1901.00279}, 2019.

\bibitem{2017generalization}
K.~Kawaguchi, L.~P. Kaelbling, and Y.~Bengio, ``Generalization in deep
  learning,'' \emph{arXiv:1710.05468}, 2017.

\bibitem{2012alexnet}
A.~Krizhevsky, I.~Sutskever, and G.~E. Hinton, ``Image{N}et classification with
  deep convolutional neural networks,'' in \emph{Adv. in Neural Inf. Process.
  Syst.}, Lake Tahoe, Nevada, Dec. 3--6, 2012, pp. 1097--1105.

\bibitem{2018deeplinear}
T.~Laurent and J.~von Brecht, ``Deep linear neural networks with arbitrary
  loss: {A}ll local minima are global,'' in \emph{Intl. Conf. on Mach. Learn.},
  Stockholm, Sweden, July 10--15, 2018, pp. 2908--2913.

\bibitem{2018multilinear}
------, ``The multilinear structure of {R}e{LU} networks,'' in \emph{Intl.
  Conf. on Mach. Learn.}, vol.~80, Stockholm, Sweden, July 10--15, 2018.

\bibitem{li2018over}
D.~Li, T.~Ding, and R.~Sun, ``Over-parameterized deep neural networks have no
  strict local minima for any continuous activations,''
  \emph{arXiv:1812.11039}, 2018.

\bibitem{li2018learning}
Y.~Li and Y.~Liang, ``Learning overparameterized neural networks via stochastic
  gradient descent on structured data,'' \emph{arXiv:1808.01204}, 2018.

\bibitem{2018adding}
S.~Liang, R.~Sun, J.~D. Lee, and R.~Srikant, ``Adding one neuron can eliminate
  all bad local minima,'' \emph{arXiv:1805.08671}, 2018.

\bibitem{1986compression}
N.~Littlestone and M.~Warmuth, ``Relating data compression and learnability,''
  University of California, Santa Cruz, Tech. Rep., 1986.

\bibitem{2018escape}
S.~Lu, M.~Hong, and Z.~Wang, ``On the sublinear convergence of randomly
  perturbed alternating gradient descent to second order stationary
  solutions,'' \emph{arXiv:1802.10418}, 2018.

\bibitem{2018separable}
M.~Nacson, N.~Srebro, and D.~Soudry, ``Stochastic gradient descent on separable
  data: {E}xact convergence with a fixed learning rate,''
  \emph{arXiv:1806.01796}, 2018.

\bibitem{2010relu}
V.~Nair and G.~E. Hinton, ``Rectified linear units improve restricted
  {B}oltzmann machines,'' in \emph{Intl. Conf. on Mach. Learn.}, Haifa, Israel,
  June 21--24, 2010, pp. 807--814.

\bibitem{2014bias}
B.~Neyshabur, R.~Tomioka, and N.~Srebro, ``In search of the real inductive
  bias: {O}n the role of implicit regularization in deep learning,''
  \emph{arXiv:1412.6614}, 2014.

\bibitem{nguyen2019connected}
Q.~Nguyen, ``On connected sublevel sets in deep learning,''
  \emph{arXiv:1901.07417}, 2019.

\bibitem{nguyen2018optimization}
Q.~Nguyen and M.~Hein, ``Optimization landscape and expressivity of deep
  {CNN}s,'' in \emph{Intl. Conf. Mach. Learn.}, 2018, pp. 3727--3736.

\bibitem{1963convergence}
A.~B. Novikoff, ``On convergence proofs for perceptrons,'' in \emph{Proc. Symp.
  Math. Theory Automata}, vol.~12, 1963, pp. 615--622.

\bibitem{2018oymak}
S.~Oymak, ``Stochastic gradient descent learns state equations with nonlinear
  activations,'' \emph{arXiv:1809.03019}, 2018.

\bibitem{oymak2019towards}
S.~Oymak and M.~Soltanolkotabi, ``Towards moderate overparameterization:
  {G}lobal convergence guarantees for training shallow neural networks,''
  \emph{arXiv:1902.04674}, 2019.

\bibitem{2016exponential}
B.~Poole, S.~Lahiri, M.~Raghu, J.~Sohl-Dickstein, and S.~Ganguli, ``Exponential
  expressivity in deep neural networks through transient chaos,'' in \emph{Adv.
  in Neural Inf. Process. Syst.}, Barcelona, Spain, Dec. 5-10, 2016, pp.
  3360--3368.

\bibitem{2017expressivity}
M.~Raghu, B.~Poole, J.~Kleinberg, S.~Ganguli, and J.~Sohl-Dickstein, ``On the
  expressive power of deep neural networks,'' in \emph{Intl. Conf. on Mach.
  Learn.}, vol.~70, Sydney, Australia, Aug. 6--11, 2017, pp. 2847--2854.

\bibitem{1958perceptron}
F.~Rosenblatt, ``The perceptron: {A} probabilistic model for information
  storage and organization in the brain,'' \emph{Psychol. Rev.}, vol.~65,
  no.~6, p. 386, Nov. 1958.

\bibitem{2018localminima}
I.~Safran and O.~Shamir, ``Spurious local minima are common in two-layer
  {R}e{LU} neural networks,'' in \emph{Intl. Conf. on Mach. Learn.}, vol.~80,
  Stockholm, Sweden, July 10--15, 2018, pp. 4430--4438.

\bibitem{book2014shai}
S.~Shalev-Shwartz and S.~Ben-David, \emph{Understanding {M}achine {L}earning:
  {F}rom {T}heory to {A}lgorithms}.\hskip 1em plus 0.5em minus 0.4em\relax New
  York, NY: Cambridge University Press, 2014.

\bibitem{2014vgg}
K.~Simonyan and A.~Zisserman, ``Very deep convolutional networks for
  large-scale image recognition,'' \emph{arXiv:1409.1556}, 2014.

\bibitem{2017learningrelus}
M.~Soltanolkotabi, ``Learning {R}e{LU} via gradient descent,'' in \emph{Adv. in
  Neural Inf. Process. Syst.}, Long Beach, CA, Dec. 4--9, 2017, pp. 2007--2017.

\bibitem{2017mahdi}
M.~Soltanolkotabi, A.~Javanmard, and J.~D. Lee, ``Theoretical insights into the
  optimization landscape of over-parameterized shallow neural networks,''
  \emph{arXiv:1707.04926}, 2017.

\bibitem{2018jlmr}
D.~Soudry, E.~Hoffer, M.~S. Nacson, S.~Gunasekar, and N.~Srebro, ``The implicit
  bias of gradient descent on separable data,'' \emph{J. Mach. Learn. Res.},
  vol.~19, no.~70, pp. 1--57, 2018.

\bibitem{1999noise}
C.~Wang and J.~C. Principe, ``Training neural networks with additive noise in
  the desired signal,'' \emph{IEEE Trans. Neural Netw.}, vol.~10, no.~6, pp.
  1511--1517, Nov. 1999.

\bibitem{taf}
G.~Wang, G.~B. Giannakis, and Y.~C. Eldar, ``Solving systems of random
  quadratic equations via truncated amplitude flow,'' \emph{IEEE Trans. Inf.
  Theory}, vol.~64, no.~2, pp. 773--794, Feb. 2018.

\bibitem{2018zhou2}
T.~Xu, Y.~Zhou, K.~Ji, and Y.~Liang, ``Convergence of {SGD} in learning
  {R}e{LU} models with separable data,'' \emph{arXiv:1806.04339}, 2018.

\bibitem{yehudai2019power}
G.~Yehudai and O.~Shamir, ``On the power and limitations of random features for
  understanding neural networks,'' \emph{arXiv:1904.00687}, 2019.

\bibitem{2018critical}
C.~Yun, S.~Sra, and A.~Jadbabaie, ``A critical view of global optimality in
  deep learning,'' \emph{arXiv:1802.03487}, 2018.

\bibitem{yun2018efficiently}
------, ``Efficiently testing local optimality and escaping saddles for {ReLU}
  networks,'' \emph{arXiv:1809.10858}, 2018.

\bibitem{yun2018small}
------, ``Small nonlinearities in activation functions create bad local minima
  in neural networks,'' \emph{arXiv:1802.03487}, 2018.

\bibitem{zwg2018arxiv}
L.~Zhang, G.~Wang, and G.~B. Giannakis, ``Real-time power system state
  estimation and forecasting via deep neural networks,''
  \emph{arXiv:1811.06146}, Nov. 2018.

\bibitem{2018gu}
X.~Zhang, Y.~Yu, L.~Wang, and Q.~Gu, ``Learning one-hidden-layer {R}e{LU}
  networks via gradient descent,'' \emph{arXiv:1806.07808}, 2018.

\bibitem{2017zhong}
K.~Zhong, Z.~Song, P.~Jain, P.~L. Bartlett, and I.~S. Dhillon, ``Recovery
  guarantees for one-hidden-layer neural networks,'' in \emph{Intl. Conf. on
  Mach. Learn.}, vol.~70, Sydney, Australia, Aug. 6--11, 2017, pp. 4140--4149.

\bibitem{2018zhou}
Y.~Zhou and Y.~Liang, ``Critical points of linear neural networks: {A}nalytical
  forms and landscape properties,'' in \emph{Intl. Conf. on Learn. Rep.},
  Vancouver, BC, Canada, Apr. 30-May 3, 2018.

\end{thebibliography}

\appendix
\renewcommand{\thesection}{Appendix A}

\subsection{Proof of Theorem \ref{th:main}}
\label{proofoftheorem}

Consider Algorithm \ref{alg:sgdn} has performed $t>0$ non-zero updates with 
a sufficiently large noise variance $\gamma^2$. Observe that if all data $(\bm{x}_{i_t},y_{i_t})\in \mathcal{S}$ lead to zero update after a succession of say e.g., $pn>0$ iterations, then Algorithm \ref{alg:sgdn} has reached a global minimum with high probability as per Proposition \ref{prop:prob}.
Let $\text{vec}(\bm{W}^\top):=[\bm{w}_1^\top~ \bm{w}_2^\top~\cdots~\bm{w}_k^\top]^\top\in\mathbb{R}^{nk\times 1}$, and define
\begin{equation}\label{eq:Omega}
	\bm{\Omega}^\star:=\frac{1}{v_{\min}}\text{sgn}(\bm{v})\otimes {\bm{\omega}^\ast}^\top\in\mathbb{R}^{k\times n} 	
\end{equation}
which is constructed from the optimum $\bm{\omega}^\ast$ of the linear classifier, where $v_{\min}:=\min_{1\le j \le k} |v_j|>0$. Using the definition of $\bm{\Omega}^\ast$ and \eqref{eq:opt}, it holds that
\begin{align}
	L_{\mathcal{S}}(\bm{\Omega}^\ast)=\frac{1}{n}\sum_{i=1}^n\max\!\bigg\{0,1-y_i \sum_{j=1}^kv_j\sigma\!\left(\text{sgn}(v_j)\,{ \bm{\omega}^\ast}^\top \bm{x}_i\right) \bigg\}.\label{eq:app11}
\end{align}
For $i\in\mathcal{N}_y^+$, we clearly have $y=1$, and $\bm{\omega}^\ast\bm{x}_i>0$. In addition, for $j\in\mathcal{N}_v^+$, we find $v_j>0$ and hence ${\rm sgn}(v_j)\,{\bm{\omega}^\ast}^\top\bm{x}_i>0$, which implies $\sigma({\rm sgn}(v_j)\,{\bm{\omega}^\ast}^\top\bm{x}_i)={\bm{\omega}^\ast}^\top\bm{x}_i$; similarly, for $j\in\mathcal{N}_v^-$, we find $v_j<0$, and thus ${\rm sgn}(v_j)\,{\bm{\omega}^\ast}^\top\bm{x}_i<0$, which yields $\sigma({\rm sgn}(v_j)\,{\bm{\omega}^\ast}^\top\bm{x}_i)=0$. These considerations show that for $i\in\mathcal{N}_y^+$ in \eqref{eq:app11}, only summands with $j\in\mathcal{N}_v^+$ survive; and arguing along the same lines, we deduce that for $i\in\mathcal{N}_y^-$, only summands with $j\in\mathcal{N}_v^-$ should be present. All in all, \eqref{eq:app11} reduces to  
\begin{align}
	L_{\mathcal{S}}(\bm{\Omega}^\ast)
	&=\frac{1}{n}\sum_{i\in\mathcal{N}_y^+}\max\bigg\{0,1-
	\sum_{j\in\mathcal{N}_v^+}	\frac{v_j}{v_{\min}} y_i
	{ \bm{\omega}^\ast}^\top \bm{x}_i \bigg\} \nonumber\\
	\label{eq:app12}
	& +\frac{1}{n}\sum_{i\in\mathcal{N}_y^-}\max\!\bigg\{0,1+\sum_{j\in\mathcal{N}_v^-} \frac{v_j}{v_{\min}}  y_i{ \bm{\omega}^\ast}^\top \bm{x}_i\bigg\}.
\end{align}   
But since $y_i\,{\bm{\omega}^\ast}^\top\bm{x}_i\ge 1$, and 
$\sum_{v_j>0} v_j/v_{\min}\ge 1 $ as well as  $\sum_{v_j<0} v_j/v_{\min}\le - 1 $, we infer that $	L_{\mathcal{S}}(\bm{\Omega}^\ast)=0$, and hence $\bm{\Omega}^\ast$ is indeed a global minimum of $L_{\mathcal{S}}(\bm{W})$. 


The subsequent analysis builds critically on the following two functions (cf. \eqref{eq:Omega})
\begin{subequations}\label{eq:func}
	\begin{align}
		\phi(\bm{W}^t)&
		:=
		=	\left\langle\text{vec}\!\left({\bm{W}^t}^\top\right),\text{vec}\!\left({\bm{\Omega}^\ast}^\top\right)  \right\rangle
		\label{eq:inner}
		\\
		\psi(\bm{W}^t)&:=
		\left\|\text{vec}\!\left({\bm{W}^t}^\top\right) \right\|_2=\bigg( \sum_{j=1}^k \left\|\bm{w}_j^t\right\|_2^2 \bigg)^{1/2}
		\label{eq:norm}
	\end{align}
\end{subequations}

Using the Cauchy-Schwartz inequality, we can write
\begin{equation}\label{eq:cs}
	\frac{\left|\phi(\bm{W}^t)\right|}{\psi(\bm{W}^t)\psi(\bm{\Omega}^\ast)}=\frac{\left|\left\langle\text{vec}\!\left({\bm{W}^t}^\top\right),\text{vec}\!\left({\bm{\Omega}^\ast}^\top\right)  \right\rangle\right|}{\left\|\text{vec}\!\left({\bm{W}^t}^\top\right) \right\|_2\left\|\text{vec}\!\left({\bm{\Omega}^\ast}^\top\right) \right\|_2}\le 1. 
\end{equation}

We will next derive a lower and an upper bound for the numerator and denominator of \eqref{eq:cs}. 
Consider an iteration $t$ for which Algorithm \ref{alg:sgdn} admits a non-zero update, meaning that $\mathds{1}_{\{1-y_{i_t}\bm{v}^\top\sigma(\bm{W}^t\bm{x}_{i_t})> 0\}}=1$, or equivalently, $y_{i_t} \bm{v}^\top\sigma(\bm{W}^t\bm{x}_{i_t})=\sum_{j=1}^k y_{i_t}v_j \sigma( {\bm{w}_j^t}^\top\bm{x}_{i_t}) <1$. It will also come handy to rewrite \eqref{eq:sgdn} row-wise as 
\begin{equation}
	\bm{w}_j^{t+1}=\bm{w}_j^t+\eta y_{i_t}v_j  \mathds{1}_{\left\{ {\bm{w}_j^t}^\top\bm{x}_{i_t} + \epsilon_j^t\ge 0\right\}}\bm{x}_{i_t},~ j=1,\,2,\,\ldots,\,k.\label{eq:upda}
\end{equation} 
Combining \eqref{eq:upda} with \eqref{eq:norm}, we can upper bound $	\psi^2(\bm{W}^{t})$ in the denominator of \eqref{eq:cs} as
\begin{align}
	\psi^2(\bm{W}^{t+1})&=\sum_{j=1}^k \left \|\bm{w}_j^{t+1}\right\|_2^2\nonumber\\
	&\stackrel{(a)}{=}\sum_{j=1}^k\Big( \left \|\bm{w}_j^{t}\right\|_2^2+\eta^2v_j^2\|\bm{x}_{i_t}\|_2^2\mathds{1}_{\left\{ {\bm{w}_j^t}^\top\bm{x}_{i_t}+ \epsilon_j^t\ge 0\right\}}	+2\eta y_{i_t}v_j \,{\bm{w}_j^t}^\top\bm{x}_{i_t}\mathds{1}_{\left\{ {\bm{w}_j^t}^\top\bm{x}_{i_t} + \epsilon_j^t\ge 0\right\}}\Big)\nonumber\\
	&\stackrel{(b)}{\le } \sum_{j=1}^k \left \|\bm{w}_j^{t}\right\|_2^2+\eta^2\sum_{j=1}^kv_j^2+2\eta \sum_{j=1}^k y_{i_t}v_j \sigma\!\left({\bm{w}_j^t}^\top\bm{x}_{i_t}\right) \nonumber\\
	&\stackrel{(c)}{\le } \psi^2(\bm{W}^{t})+\eta^2\|\bm{v}\|_2^2+2\eta \label{eq:nrec}
\end{align}
where $(a)$ follows directly from \eqref{eq:upda} after expanding the squares; $(b)$ uses the working condition $\|\bm{x}_{i_t}\|_2\le 1$ adopted without loss of generality, as well as the fact that $\mathds{1}_{\{\cdot\}}\le 1$ holds true for any event, and also  the inequality $\sum_{j=1}^k y_{i_t}v_j {\bm{w}_j^t}^\top\bm{x}_{i_t}\mathds{1}_{\{{\bm{w}_j^t}^\top\bm{x}_{i_t}+ \epsilon_j^t\ge 0\}})
\le \sum_{j=1}^k y_{i_t}v_j \sigma({\bm{w}_j^t}^\top\bm{x}_{i_t})$ established in Lemma \ref{le:smaller} below, whose proof is postponed to Appendix \ref{proofofsmaller} for readability. Finally, $(c)$ is due to the non-zero update at iteration $t$, which implies that $\sum_{j=1}^k y_{i_t}v_j \sigma({\bm{w}_j^t}^\top\bm{x}_{i_t}) <1$.  

\begin{lemma}
	\label{le:smaller}
	For any $(\bm{x},y)\in\mathcal{S}$, and any $\{\bm{w}_j\in\mathbb{R}^d\}_{j=1}^k$ and $\bm{v}\in\mathbb{R}^k$, it holds that 
	\begin{equation}\label{eq:smaller}
		\sum_{j=1}^k yv_j \bm{w}_j^\top\bm{x}\mathds{1}_{\left\{\bm{w}_j^\top\bm{x} + \epsilon_j\ge 0\right\}}
		\le \sum_{j=1}^k yv_j \sigma\!\left( \bm{w}_j^\top\bm{x}\right) 
	\end{equation}
	if the entries of the additive noise $\bm{\epsilon}\in\mathbb{R}^k$ satisfy 
	$\epsilon_{j}\sim \mathcal{N}(0,\,\gamma^2)$, when $yv_j\ge 0$, and $\epsilon_{j}=0$, otherwise. 
\end{lemma}

Writing down \eqref{eq:nrec} for the already executed $t$ non-zero updates and by means of telescoping, we obtain
\begin{equation}
	\label{eq:normbound}
	\psi^2(\bm{W}^{t})\le \psi^2(\bm{W}^{0})+t\left(\eta^2\|\bm{v}\|_2^2+2\eta \right).
\end{equation}

We now turn to deriving a lower bound for $\phi(\bm{W}^t)$ in \eqref{eq:cs}, starting with (cf. \eqref{eq:inner} and \eqref{eq:norm}) 
\begin{align}
	\phi(\bm{W}^{t+1})&=\frac{1}{v_{\min}}\sum_{j\in \mathcal{N}_v^{+}}{ \bm{w}_{j}^{t+1}}^\top \bm{\omega}^\ast-\frac{1}{v_{\min}}\sum_{j\in\mathcal{N}_{v}^-} { \bm{w}_{j}^{t+1}}^\top \bm{\omega}^\ast \nonumber\\
	&\stackrel{(a)}{=}\frac{1}{v_{\min}}\sum_{j\in \mathcal{N}_v^+}{ \bm{w}_{j}^{t}}^\top \bm{\omega}^\ast-\frac{1}{v_{\min}}\sum_{j\in\mathcal{N}_{v}^-} { \bm{w}_{j}^{t}}^\top \bm{\omega}^\ast \nonumber\\
	&\quad	+\frac{1}{v_{\min}}\sum_{j\in \mathcal{N}_v^+}
	\eta v_j y_{i_{t}}\bm{x}_{i_{t}}^\top\bm{\omega}^\ast
	\mathds{1}_{\left\{ {\bm{w}_j^{t}}^\top\bm{x}_{i_{t}} + \epsilon_j^{t}\ge 0\right\}}
-\frac{1}{v_{\min}}\sum_{j\in \mathcal{N}_v^-}
	\eta v_j y_{i_{t}} \bm{x}_{i_{t}}^\top\bm{\omega}^\ast \mathds{1}_{\left\{{\bm{w}_j^{t}}^\top\bm{x}_{i_{t}} + \epsilon_j^{t}\ge 0\right\}}
	\nonumber\\
	&\stackrel{(b)}{=} \phi(\bm{W}^{t})+\eta
	y_{i_{t}} \bm{x}_{i_{t}}^\top\bm{\omega}^\ast \sum_{j=1}^k
	\frac{|v_j|}{v_{\min}}  \mathds{1}_{\left\{ {\bm{w}_j^{t}}^\top\bm{x}_{i_{t}} + \epsilon_j^{t}\ge 0\right\}}\nonumber\\
	&\stackrel{(c)}{\ge } \phi(\bm{W}^{t})+\eta \label{eq:irec}
\end{align}
where $(a)$ is derived by plugging in \eqref{eq:upda};  $(b)$ uses $v_j>0$ ($<0$) if $j\in\mathcal{N}_v^+$ ($\in\mathcal{N}_v^-$);  
and $(c)$ follows from the two critical inequalities: i) $y_{i} \bm{x}_{i}^\top\bm{\omega}^\ast\ge 1$ for all $(\bm{x}_i,y_i)\in\mathcal{S}$, and ii) $ (|v_j|/v_{\min}) \mathds{1}_{\{ {\bm{w}_j^{t}}^\top\bm{x}_{i_{t}} + \epsilon_j^{t}\ge 0\}}\ge 1$, because a non-zero update at iteration $t$ asserts that \emph{at least one} out of the $k$ ReLU activity indicator functions $\{ \mathds{1}_{\{ {\bm{w}_j^{t}}^\top\bm{x}_{i_{t}} + \epsilon_j^{t}\ge 0\}} \}_{j=1}^k$ equals one.  

Again, telescoping the $t$ recursions \eqref{eq:irec} for the non-zero updates $0$ to $(t-1)$, yields
\begin{equation}
	\label{eq:innerbound}
	\phi(\bm{W}^t)\ge \phi(\bm{W}^0) + t\eta.
\end{equation}

Substituting the bounds in \eqref{eq:normbound} and \eqref{eq:innerbound} into \eqref{eq:cs}, we have that
\begin{align}
	\label{eq:established}
	\phi(\bm{W}^0) + t\eta &\le	\left|\phi(\bm{W}^t)\right|
	\le \psi(\bm{W}^t) \psi(\bm{\Omega}^\ast)
	=\sqrt{\psi^2(\bm{W}^{0})+t\left(\eta^2\|\bm{v}\|_2^2+2\eta \right) }\; \psi(\bm{\Omega}^\ast).
\end{align}
Using further that $\sqrt{p^2+q^2}\le |p|+|q|$, we arrive at
\begin{align}
	\label{eq:less}
	\phi(\bm{W}^0) + t\eta &\le	\left|\phi(\bm{W}^t)\right|
	\le  \psi(\bm{W}^0) \psi(\bm{\Omega}^\ast)	+\sqrt{t} \cdot\sqrt{\eta^2\left\|\bm{v}\right\|_2^2+2\eta} \;\psi(\bm{\Omega}^\ast).
\end{align}

Using that $[{\rm sgn}(v_j)]^2=1$, it is easy to verify that
\begin{align}\label{eq:bound1}
	\psi(\bm{\Omega}^\ast)&:=\left\|\text{vec}\!\left({\bm{\Omega}^\ast}^\top\right)\right\|_2=\left\|\bm{\Omega}^\ast\right\|_{F}=\frac{1}{v_{\min}}\left\|\text{sgn}(\bm{v})\otimes {\bm{\omega}^\ast}^\top\right\|_F=\frac{\sqrt{k}}{v_{\min}}\|\bm{\omega}^\ast\|_2.
\end{align}
Under our assumption that all rows of $\bm{W}^0$ satisfy $\|\bm{w}_j^0\|_2\le \rho$, we have for $\psi(\bm{W}^0):=\|\bm{W}^0\|_{F}$ that 
\begin{equation}\label{eq:bound2}
	\psi(\bm{W}^0)\le \sqrt{k}\rho
\end{equation}

Using \eqref{eq:inner} along with \eqref{eq:bound1} and \eqref{eq:bound2}, we find
\begin{align}
	\phi(\bm{W}^0)&=\left\langle\text{vec}\!\left({\bm{W}^0}^\top\right),\text{vec}\!\left({\bm{\Omega}^\ast}^\top\right)\right\rangle \nonumber\\	&\ge -\left\|\text{vec}\!\left({\bm{W}^0}^\top\right)\right\|_2 
	\left\|\text{vec}\!\left({\bm{\Omega}^\ast}^\top\right)\right\|_2\nonumber\\
	&=-\psi(\bm{W}^0)\psi(\bm{\Omega}^\ast)\nonumber\\
	&\ge -\frac{k\rho}{v_{\min}}\left\|\bm{\omega}^\ast\right\|_2.\label{eq:bound3}
\end{align}
Substituting the bounds in \eqref{eq:bound1}, \eqref{eq:bound2}, and \eqref{eq:bound3} into \eqref{eq:less} and re-arranging terms, we further arrive at
\begin{equation}
	\eta v_{\min} t\le \|\bm{\omega}^\ast\|_2 \sqrt{k\left(\eta^2\left\|\bm{v}\right\|_2^2+2\eta\right)} \sqrt{t}+ 2k\rho\left\|\bm{\omega}^\ast\right\|_2
\end{equation}
which upon letting $z:=\sqrt{t}\ge 0$, boils down to the quadratic inequality
\begin{equation}
	a z^2+b z+c\le 0\qquad {\rm s.\,to}\qquad z\ge 0 \label{eq:quad}
\end{equation}
where the coefficients are given by $a=\eta v_{\min}>0$, $b= -\|\bm{\omega}^\ast\|_2 \sqrt{k(\eta^2\|\bm{v}\|_2^2+2\eta)}$, and $c=-2k\rho\|\bm{\omega}^\ast\|_2<0$. Because $c<0$ and $b^2-4ac> 0$, we have real roots of opposite sign, which implies that \eqref{eq:quad} is satisfied for 
\begin{equation}
	z\in \left[0,\; \frac{-b+\sqrt{b^2-4ac}}{2a}\right].
\end{equation}
Plugging in those coefficients and appealing again to the inequality $\sqrt{p^2+q^2}\le |p|+|q|$, we deduce that
\begin{align}
	t=z^2&\le \frac{b^2+b^2-4ac-2b\sqrt{b^2-4ac}}{4a^2}\nonumber\\
	&\le \frac{b^2}{2a^2}-\frac{c}{a}+\frac{b^2}{2a^2}-\frac{b\sqrt{-ac}}{a^2} 
	\nonumber\\
	&\le \frac{ k\left(\eta^2\left\|\bm{v}\right\|_2^2+2\eta\right)\left\|\bm{\omega}^\ast\right\|_2^2}{2\eta^2v_{\min}^2}+\frac{2k\rho\eta v_{\min} \left\|\bm{\omega}^\ast\right\|_2^2}{\eta^2v_{\min}^2}\nonumber\\
	&\quad 	+\frac{k\left(\eta^2\left\|\bm{v}\right\|_2^2+2\eta\right)\left\|\bm{\omega}^\ast\right\|_2^2 }{2\eta^2v_{\min}^2}+\frac{\left\|\bm{\omega}^\ast\right\|_2 \sqrt{k\left(\eta^2\left\|\bm{v}\right\|_2^2+2\eta\right)}\sqrt{2k\rho\eta v_{\min} \left\|\bm{\omega}^\ast\right\|_2}}{\eta^2v_{\min}^2}\nonumber\\
	&=\frac{k}{\eta v_{\min}^2}\Big[\left(\eta\left\|\bm{v}\right\|_2^2+2\right)\left\|\bm{\omega}^\ast\right\|_2^2+2\rho v_{\min}\left\|\bm{v}\right\|_2^2
	+\sqrt{2 \rho v_{\min} \left\|\bm{\omega}^\ast\right\|_2\left( \eta\left\|\bm{v}\right\|_2^2+2\right)}\left\|\bm{\omega}^\ast\right\|_2
	\Big]\nonumber\\
	&\stackrel{\triangle}{=}T_k.\label{eq:tk}
\end{align}  

By taking $\rho=0$ in \eqref{eq:tk}, one finally confirms that the maximum number of non-zero updates for Algorithm \ref{alg:sgdn} initialized with $\bm{W}^0=\bm{0}$ until convergence, is
\begin{equation}
	T_k^0:=\frac{k}{\eta v_{\min}^2}\left(\eta\left\|\bm{v}\right\|_2^2+2\right)\!\left\|\bm{\omega}^\ast\right\|_2^2
\end{equation}
which completes the proof.

\renewcommand{\thesection}{Appendix B}
\subsection{Proof of Lemma \ref{le:smaller}}\label{proofofsmaller}


We first prove that the following inequality holds per hidden neuron $j= 1,\,2,\,\ldots,\,k$
\begin{equation}\label{eq:cons}
	yv_j \bm{w}_j^\top\bm{x}\mathds{1}_{\left\{\bm{w}_j^\top\bm{x} + \epsilon_j\ge 0\right\}}\le yv_j \sigma\!\left( \bm{w}_j^\top\bm{x}\right) .
\end{equation}
Depending on whether the $j$-th ReLU is active or not ($\bm{w}_j^\top\bm{x} \gtreqless 0 $)
and (Gaussian) noise is injected or not ($v_jy\gtreqless 0$), 
we consider separately the following four cases:
\begin{enumerate}
	\item [c1)] $\bm{w}_j^\top\bm{x}\ge 0$ and $v_j y\ge 0$ (ReLU active and noise injected); 
	\item [c2)] $\bm{w}_j^\top\bm{x}\ge 0$ and $v_j y< 0$ (ReLU active and no noise);
	\item [c3)] $\bm{w}_j^\top\bm{x}< 0$ and $v_j y\ge 0$ (ReLU inactive and noise injected); and, 
	\item [c4)] $\bm{w}_j^\top\bm{x}< 0$ and $v_j y< 0$ (ReLU inactive and no noise). 
\end{enumerate}

For c1), the right-hand-side of \eqref{eq:cons} satisfies 
\begin{equation}\label{eq:appc1r}
	yv_j \sigma(\bm{w}_j^\top\bm{x})
	=yv_j\bm{w}_j^\top\bm{x}.
\end{equation}
Regarding the left-hand-side, it takes values depending on whether $\epsilon_j$ changes the state of the ReLU activity indicator function, it leads to a two-branch inequality 
\begin{align}\label{eq:appc1l}
	yv_j \bm{w}_j^\top\bm{x}\mathds{1}_{\left\{\bm{w}_j^\top\bm{x} + \epsilon_j\ge 0\right\}}
	=\left\{\begin{array}
		{lc}
		yv_j \bm{w}_j^\top\bm{x}, &  \epsilon_j\ge -\bm{w}_j^\top\bm{x}\\
		0,&\epsilon_j< -\bm{w}_j^\top\bm{x}
	\end{array}
	\right..
\end{align}
Combining \eqref{eq:appc1r} with \eqref{eq:appc1l}, we deduce that $ yv_j \bm{w}_j^\top\bm{x}\mathds{1}_{\{\bm{w}_j^\top\bm{x} + \epsilon_j\ge 0\}}\le yv_j \sigma(\bm{w}_j^\top\bm{x}) $ holds under c1). 

For c2),  the $j$-th ReLU is active too, but there is no noise injection, namely $\epsilon_j=0$. The right-hand-side of \eqref{eq:appc1r} still holds however. It is also not difficult to check the left-hand-side term 
$yv_j \bm{w}_j^\top\bm{x}\mathds{1}_{\{\bm{w}_j^\top\bm{x} + \epsilon_j\ge 0\}}=yv_j \bm{w}_j^\top\bm{x}$. Evidently, the desired inequality holds with equality in this case.

For c3), we have 
$\bm{w}_j^\top\bm{x}< 0$, meaning that the $j$-th ReLU is inactive, and therefore, the right-hand-side of \eqref{eq:cons} becomes $yv_j \sigma(\bm{w}_j^\top\bm{x})=0$. However, given $v_j y\ge 0$, there is a noise injection. Hence, the left-hand-side can be similarly treated as in c2), to infer that \eqref{eq:appc1l} remains valid. Recalling again that $\bm{w}_j^\top\bm{x}< 0$ and $yv_j\ge 0$, one deduces that $yv_j \bm{w}_j^\top\bm{x}\mathds{1}_{\{\bm{w}_j^\top\bm{x} + \epsilon_j\ge 0\}}\le 0 $, regardless of $\epsilon_j$. Thus, the inequality under consideration is also true under c2).

Finally, for c4), the $j$-th ReLU is inactive, and there is no noise injection. It is straightforward to verify that both the left-hand-side and right-hand-side equal zero, and \eqref{eq:cons} holds with equality as well. 

Putting together c1)-c4), we have established that $ yv_j \bm{w}_j^\top\bm{x}\mathds{1}_{\{\bm{w}_j^\top\bm{x} + \epsilon_j\ge 0\}}\le yv_j \sigma( \bm{w}_j^\top\bm{x}) $ for $j=1,\,2,\,\ldots,\,k$. Summing up such inequalities for all $k$ hidden neurons completes the proof.



\renewcommand{\thesection}{Appendix C}
\subsection{Proof of Theorem \ref{th:lower}}\label{proofoflower}

Let $\{\bm{e}_i\in\mathbb{R}^d\}_{i=1}^d$ be the canonical basis of $\mathbb{R}^d$. 
Consider the following set $\mathcal{S}_1\subseteq \mathcal{X}\times \mathcal{Y}$ of $d$ training data from the `positive' class 
\begin{equation}
	\label{eq:set}
	\mathcal{S}_1:=\left\{\left(\bm{e}_1,1\right),\left(\bm{e}_2,1\right),\ldots,\left(\bm{e}_d,1\right)\right\}.
\end{equation}
Letting $\bm{\omega}^\ast:=[1~1~\cdots~1]^\top\in\mathbb{R}^d$, it is clear that the linear classifier $y=\bm{w}_j^\top\bm{x} $ correctly classifies all the $d$ data points in $\mathcal{S}$. Note also that  $\|\bm{\omega}^\ast\|_2^2=d$ in this case. 

Consider the update in \eqref{eq:sgdn} initialized with $\bm{W}^0=\bm{0}$, and telescope each row to obtain
\begin{align}
	\bm{w}_j^{t+1}=\eta v_j \sum_{i=1}^{t} \mathds{1}_{\left\{ {\bm{w}_j^t}^\top\bm{e}_{i_t} + \epsilon_j^t\ge 0\right\}}\bm{e}_{i_t},\quad  j=1,\,2,\,\ldots,\,k\label{eq:updanew}
\end{align}
where $i_t$ deterministically cycles through $\{1,\,2,\,\ldots,\,d\}$.

At the global optimum, say $\bm{W}^{\tau}$ for some iteration number $\tau$, it holds for $(\bm{e}_s,1 )\in\mathcal{S}_1$ that 
\begin{align}
	y_s f(\bm{e}_s;\bm{W})&=\sum_{j=1}^k v_j\sigma\!\left({\bm{w}_j^{\tau}}^\top\bm{e}_s\right)=\sum_{j\in\mathcal{N}_v^+}v_j\sigma\!\left({\bm{w}_j^{\tau}}^\top\bm{e}_s\right)-\sum_{j\in\mathcal{N}_v^-}|v_j|\sigma\!\left({\bm{w}_j^{\tau}}^\top\bm{e}_s\right)\ge 1.\label{eq:above}
\end{align}
Since $|v_j|\sigma({\bm{w}_j^\tau}^\top\bm{e})\ge 0$ in \eqref{eq:above}, 
a necessary condition for $\bm{W}^\tau$ to be a global minimum is (cf. \eqref{eq:updanew})
\begin{align}\label{eq:positive}
	1&\le 	\sum_{j\in\mathcal{N}_v^+}v_j\sigma\!\left({\bm{w}_j^{\tau-1}}^\top\bm{e}_s\right)=\sum_{j\in\mathcal{N}_v^+}v_j\left\langle\bm{w}_j^{\tau},\bm{e}_s\right\rangle\nonumber\\
	&=\sum_{j\in\mathcal{N}_v^+}v_j\left\langle\eta v_j \sum_{t=1}^{\tau-1} \mathds{1}_{\left\{ {\bm{w}_j^t}^\top\bm{e}_{i_t} + \epsilon_j^t\ge 0\right\}}\bm{e}_{i_t},\bm{e}_s\right\rangle\nonumber\\
	&\ge 
	1.
\end{align}

Assume for simplicity that $\tau-1$ is a multiple of $d$, namely $\tau-1=dp$ for some integer $p\ge 1$. On the other hand, we have from \eqref{eq:positive} that
\begin{align}
		\sum_{j\in\mathcal{N}_v^+}v_j\left\langle\eta v_j \sum_{i=1}^{\tau-1} \mathds{1}_{\left\{ {\bm{w}_j^t}^\top\bm{e}_{i_t} + \epsilon_j^t\ge 0\right\}}\bm{e}_{i_t},\bm{e}_s\right\rangle 
	&\le  	\sum_{j\in\mathcal{N}_v^+}\eta v_j^2 \left\langle\sum_{i=1}^{\tau-1} \bm{e}_{i_t},\bm{e}_s\right\rangle
=\eta p \sum_{j\in\mathcal{N}_v^+} v_j^2\le p\eta \|\bm{v}\|_2^2\label{eq:positive1}
\end{align}
where we have used the following inequalities: i) $\mathds{1}_{\left\{{\bm{w}_j^t}^\top\bm{e}_{i_t} + \epsilon_j^t\ge 0\right\}}\le 1$, ii) $\sum_{i=1}^{\tau-1} \bm{e}_{i_t}=\sum_{i=1}^p \bm{1}$ with $\bm{1}$ being an all-one vector of suitable dimension that is clear from the context, 
and iii) $\sum_{j\in\mathcal{N}_v^+}v_j^2\le \|\bm{v}\|_2^2$.

Combing the bounds in \eqref{eq:positive} and \eqref{eq:positive1}, we obtain that $p\eta \|\bm{v}\|_2^2\ge 1$, or equivalently $p\ge 1/(\eta \|\bm{v}\|_2^2)$. Hence, to find a global optimum, Algorithm \ref{alg:sgdn} initialized from $\bm{W}^0=\bm{0}$ makes at least 
\begin{equation}\label{eq:lower}
	M_k^0 \ge \frac{d}{\eta \|\bm{v}\|_2^2}=\frac{\|\bm{\omega}^\ast\|_2^2}{\eta \|\bm{v}\|_2^2}
\end{equation}
mistakes. This concludes the proof.

\renewcommand{\thesection}{Appendix D}

\subsection{Proof of Proposition \ref{prop:prob}}\label{proofofprop}

We have established in \eqref{eq:normbound} that
\begin{equation}
	\label{eq:normbound1}
	\psi^2(\bm{W}^{t})=\sum_{j=1}^k \left \|\bm{w}_j^t \right\|_2^2  \le \psi^2(\bm{W}^{0})+t\left(\eta^2\|\bm{v}\|_2^2+2\eta \right).
\end{equation}
We have also proved in Theorem \ref{th:main} that Algorithm \ref{alg:sgdn} performs at most $T_{k}$ non-zero updates regardless of $\gamma^2$ (cf. \eqref{eq:tk}). Hence, as long as the initialization $\bm{W}^0$ is bounded, all iterates $\bm{W}^t$ will be bounded; that is, there exists some constant $w_{\max}>0$ such that $\|\bm{w}_j^t\|_2\le w_{\max} $ holds for all $j=1,\,2,\,\ldots,\,k$ and $t=0,\,1,\,\ldots,\,T_k$.

For notational brevity, we drop the iteration index $t$,  and let the current iterate be denoted by $\bm{W}$. If there is no non-zero update with the current sampled data point $(\bm{x}_{i},y_{i} )\in\mathcal{S}$, then one of the following two cases must be true: c1) $\mathds{1}_{\{1- y_{i}\bm{v}^\top\sigma(\bm{W}\bm{x}_{i}) >0\}}=0$, or equivalently, $\max(0,1- y_{i}\bm{v}^\top\sigma(\bm{W}\bm{x}_{i}))=0$ implying that $(\bm{x}_{i},y_{i} )$ is correctly classified; and, c2) $\mathds{1}_{\{\bm{W}\bm{x}_{i}+\bm{\epsilon}\ge \bm{0} \} }=\bm{0} $, or equivalently, $\epsilon_j<- \bm{w}_j^\top\bm{x}_{i}$ for all neurons $j=1,\,2,\,\ldots,\,k$.  

When $i$ cycles through $\{1,\,2,\,\ldots,\,n \}$ in a deterministic manner (with each integer drawn exactly once every $n$ iterations), then within every succession of $np$ iterations, the noise-injected stochastic gradient term $\mathds{1}_{\{1-y_{i}\bm{v}^\top\sigma(\bm{W}\bm{x}_{i})> 0\}}\cdot  y_{i}\bm{v}\, \text{diag}(\mathds{1}_{\{\bm{W}\bm{x}_{i}+\bm{\epsilon}\ge \bm{0}\}})\,\bm{x}_{i}^\top$ in \eqref{eq:sgdn} will be evaluated exactly $p$ times at every datum $(\bm{x}_i,y_i )\in\mathcal{S}$, but with $p$ different random noise realizations. Hence, if there is no non-zero update within a succession of $np$ iterations, the probability of event c2) occurring $np$ times is at most  
\begin{align} 
	\left[ \prod_{j \in \min\left\{ \mathcal{N}_v^+,\,\mathcal{N}_v^-\right\} }
	\Phi\!\left(\frac{\max_{1\le i\le n } - \bm{w}_j^\top \bm{x}_i  }{\gamma}
	\right) 
	\right]^p\le 
	\left[
	\Phi\!\left(\frac{w_{\max} }{\gamma}
	\right) 
	\right]^{p\min\left\{ \left|\mathcal{N}_v^+\right|, \left|\mathcal{N}_v^-\right|\right\}}\label{eq:46}
\end{align}
when there is only one out of the $n$ data points that is left incorrectly classified.
Here, by a slight abuse of notation, we use $j \in \min\{ \mathcal{N}_v^+,\,\mathcal{N}_v^-\}$ to mean $j\in\mathcal{N}_v^+$ if $|\mathcal{N}_v^+|\le |\mathcal{N}_v^-|$, and $j\in\mathcal{N}_v^-$ otherwise. Furthermore, to obtain the inequality in \eqref{eq:46}, we have used $\max_{1\le i\le n } (- \bm{w}_j^\top \bm{x}_i )\le \|\bm{w}_j\|_2\|\bm{x}_i\|_2 \le w_{\max}$ under our assumptions that $ \|\bm{w}_j\|_2\le w_{\max}$ and $\|\bm{x}_i\|_2\le 1$ for all $j=1,\,2,\,\ldots,\,k$ and $i=1,\,2,\,\ldots,\,n$.
Therefore, by the total probability theorem, the probability of having c1) hold for all data points  
is at least
\begin{equation} 
	1-
	\left[
	\Phi\!\left(\frac{w_{\max} }{\gamma}
	\right) 
	\right]^{p\min\left\{ \left|\mathcal{N}_v^+\right|, \left|\mathcal{N}_v^-\right|\right\}}
\end{equation}
which can be made arbitrarily close to $1$ by taking either a large enough $p$ and/or $\gamma>0$. 
The case of picking $i_t$ uniformly at random from $\{1,\,2,\,\ldots,\,n \}$ can be discussed in a similar fashion, but it is omitted here. This completes the proof.

\end{document}